\documentclass[lettersize,journal]{IEEEtran}
\usepackage{amsmath,amsfonts}
\usepackage{algorithm}
\usepackage{array}
\usepackage[caption=false,font=normalsize,labelfont=sf,textfont=sf]{subfig}
\usepackage{textcomp}
\usepackage{stfloats}
\usepackage{url}
\usepackage{verbatim}
\usepackage{graphicx}
\usepackage{cite}
\usepackage{algpseudocode}
\usepackage{makecell}
\usepackage{multirow}
\usepackage{amssymb}
\usepackage{booktabs}
\usepackage[table]{xcolor}

\begin{document}

\title{Low-rank Orthogonal Subspace Intervention for Generalizable Face Forgery Detection}

\author{Chi Wang, Xinjue Hu, Boyu Wang, Ziwen He,~\IEEEmembership{Member,~IEEE}, Zhangjie Fu,~\IEEEmembership{Member,~IEEE}
\thanks{This work was supported in part by the National Natural Science Foundation of China under grant U22B2062, 62172232, 62402228, 62172234, by the Jiangsu Provincial Science and Technology Major Project under grant BG2024042, and by the Natural Science Foundation of Jiangsu Province under grant BK20240699. (Chi Wang and Xinjue Hu contributed equally to this
work.) (Corresponding author: Zhangjie Fu).}
\thanks{Chi Wang, Xinjue Hu, Boyu Wang, Ziwen He and Zhangjie Fu are with the Engineering Research Center of Digital Forensics, Ministry of Education, Nanjing University of Information Science and Technology, Nanjing, 210044, China (e-mail: 202412200715@nuist.edu.cn; xinjueh@126.com; 202412200714@nusit.edu.cn; ziwen.he@nuist.edu.cn; fzj@nuist.edu.cn).}
}

\markboth{Journal of \LaTeX\ Class Files,~Vol.~14, No.~8, August~2021}%
{Shell \MakeLowercase{\textit{et al.}}: A Sample Article Using IEEEtran.cls for IEEE Journals}

\IEEEpubid{0000--0000/00\$00.00~\copyright~2021 IEEE}

\maketitle

\begin{abstract}
The generalization problem remains a key challenge in face forgery detection. This paper explores the reasons for the generalization failure of Vanilla CLIP: in ``real vs. fake" detection, the few dominant principal components in the feature space primarily encode forgery-irrelevant information, rather than authentic forgery traces. However, this irrelevant information inevitably leads to spurious correlations, severely limiting detector performance. We define this phenomenon as ``\textit{low-rank spurious bias}". To address this, we propose a low-rank representation space intervention paradigm, named the SeLop, from the perspective of causal representation learning. SeLop unifies the spurious correlation factors irrelevant to forgery into a low-rank subspace and cuts off the statistical shortcut between it and the label, thus aligning representation learning with authentic forgery traces. Specifically, we decompose spurious correlation features into a low-rank subspace through orthogonal low-rank projection, then remove this subspace from the original representation and train its orthogonal complement to capture forgery-related features. This low-rank projection removal effectively eliminates spurious correlation factors, ensuring that classification decisions are based on authentic forgery cues. With only 0.39M trainable parameters, our method achieves state-of-the-art performance across several benchmarks, demonstrating excellent robustness and generalization.
\end{abstract}

\begin{IEEEkeywords}
Face Forgery Detection, CLIP, Causal Representation Learning, Generalization.
\end{IEEEkeywords}

\section{Introduction}
\IEEEPARstart{T}{hanks} to the rapid development of artificial intelligence, face forgery techniques\footnote{Face forgery strictly refers to techniques that manipulate local facial content, e.g., lip motion editing, central-face swapping, etc. Full-image synthesis via GAN/Diffusion is out of scope.} has made it possible to easily manipulate a real face (e.g., face swapping \cite{faceswap}, expression modification \cite{beijing1}). In recent years, facial manipulation has attracted considerable attention due to its convenience and has been widely used in film and television production. However, if this technology is used maliciously (e.g., digital fraud and spread of fake news), it will undoubtedly pose a significant threat to social security and media credibility \cite{yaoqiu1,yaoqiu2}. Therefore, there is an urgent need to develop a generalizable and robust face forgery detector.
\IEEEpubidadjcol

\begin{figure}[t]
  \centering
  \includegraphics[width=\columnwidth]{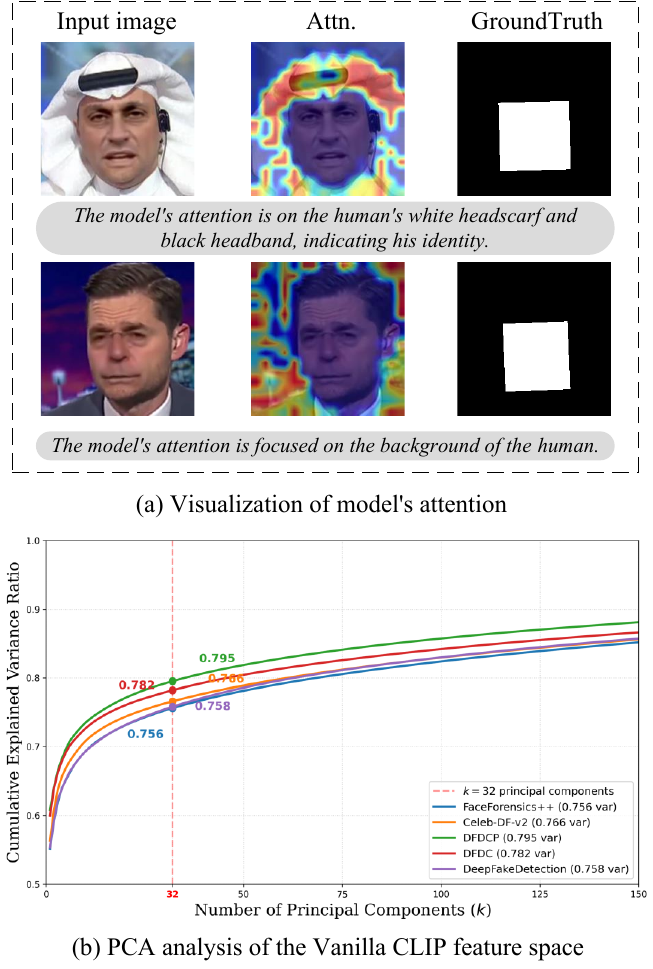}
  \caption{(a) Visualization of Vanilla CLIP attention mapping. We can observe that \textbf{Vanilla CLIP always pays attention to forgery-irrelevant regions} (e.g., identity and background) and treats them as discriminative cues. (b) PCA analysis for Vanilla CLIP's feature space across multiple deepfake datasets. The figures show that CLIP's 1024-dimensional high-dimensional feature space generally exhibits a \textbf{low-rank manifold distribution} across the datasets: the first 32 principal components alone explain over 75\% of the total feature variance.}
  \label{fig:motivation}
\end{figure}
\IEEEpubidadjcol
Early paradigms typically define face forgery detection as a standard ``real vs fake" binary classification problem. They utilize deep neural networks to extract facial features, which are then fed into a classifier to obtain the probability of ``fake" for classification. In addition, some works have attempted to utilize different auxiliary information to mine forgery clues, including frequency signals \cite{f2trans,f3}, biological signals \cite{lips,gaze}, spatiotemporal clues \cite{fcg}, and so on. Although they achieve good results on the training set, their performance drops greatly when facing unknown face manipulation methods. A common explanation \cite{sbi} is: these methods often fall into local optimal solutions and overfit specific forgery patterns in training data, which hinders their performance in cross-dataset evaluation.

A practical remedy is to leverage pre-trained knowledge from visual foundation models (VFMs) to expand the feature space and mitigate overfitting \cite{orthogonal}. CLIP \cite{clip}, a representative vision–language model, demonstrates strong zero-shot transfer on downstream tasks. Motivated by this, recent studies have adapted CLIP for face forgery detection using two main routes: adapter-based structured finetuning \cite{fcg,FA,Moeffd} and prompt-based conditional alignment \cite{mfclip, gaints}. Empirical results show that pre-training knowledge helps capture forgery cues and yields notable improvements. However, CLIP’s semantic space is large and structurally complex. Forgery-related knowledge and forgery-irrelevant factors are often entangled within the same representation. When relying solely on external adapters or prompts, there is no explicit mechanism to decompose and constrain these two components inside the representation space. As a result, optimization can absorb task-irrelevant statistical patterns, which increases sensitivity to domain shifts and reduces cross-distribution stability. Guided by these observations, we pose the core problem as follows: How can CLIP’s internal representations be adjusted during training so that the model focuses on modeling forgery cues while suppressing forgery-irrelevant factors?

To delve into this problem, we first visualize the attention of Vanilla CLIP\cite{clip} using GradCAM\cite{cam}, as shown in Fig. \ref{fig:motivation} (a). Untuned CLIP tends to focus on forgery-irrelevant information (e.g., human's identity and background), treating them as discriminative cues. This is because CLIP was pretrained on 400 million image-text pairs and did not directly learn the meaning of ``forgery", resulting in strong semantic prioritization and image semantic generalization capabilities. Furthermore, we conduct an empirical PCA study on the Vanilla CLIP representation space on five deepfake datasets, as shown in Fig. \ref{fig:motivation} (b). We randomly sample 100,000 patch features from each dataset for PCA analysis and calculate the cumulative explained variance ratio. A striking and consistent phenomenon emerges: the feature manifold exhibits a ``low-rank" distribution, with a very small number of principal components explaining the vast majority of the variance (e.g., over 75\% of the total variance is concentrated in the 32 principal components). This indicates that the Vanilla CLIP's features are highly redundant. The few principal components that contribute significantly are essentially fitting natural semantic variations in real-world images, rather than real forgery traces (combined with Fig. \ref{fig:motivation} (a) and Fig. \ref{fig:motivation} (b)). However, for face forgery detection, these inherent biases often lead the model to take shortcuts and generate spurious correlations. Simultaneously, subtle forgery traces are severely suppressed and ``drown" in a flat, long-tailed subspace, making standard fine-tuning difficult to capture robust, generalizable patterns. We define the above phenomenon as \textit{\textbf{low-rank spurious bias}}.

To address this, we propose a novel approach called \textbf{SeLop} (\textbf{S}purious correlation \textbf{e}limination via \textbf{L}ow-rank \textbf{o}rthogonal \textbf{p}rojection for face forgery detection). SeLop innovatively addresses the \textit{low-rank spurious bias} phenomenon from the perspective of causal representation learning. We define forgery-irrelevant information that causes spurious correlation as spurious correlation factors, while the authentic forgery traces are causal features. SeLop's core idea is to treat these spurious correlation factors as a whole, performing a unified subspace-level intervention in the CLIP visual representation space and cutting off the backdoor path from spurious correlation factors to groundtruth labels. Specifically, given the low-rank manifold distribution of the original CLIP feature space on deepfake data, we introduce a trainable basis matrix and construct a compact orthogonal subspace through QR decomposition. Subsequently, we explicitly remove this low-rank subspace from the original representation through orthogonal projection operations, forcing the model to mine sparse but essential causal forgery cues in the remaining orthogonal complement space. This mechanism achieves a dataset-independent unified elimination. Through end-to-end training with only \textbf{0.39M} trainable parameters, the model is able to learn causal features related to forgery while preserving the rich pre-trained knowledge of CLIP to the maximum extent.

Overall, the main contributions of this paper are as follows:
\begin{enumerate}
\item{\textit{Low-rank spurious bias} phenomenon in Vanilla CLIP: Through GradCAM visualization of the attention map and PCA analysis for feature space's energy spectrum, we discover \textit{low-rank spurious bias}. That is, a few significantly contributing principal components encode forgery-irrelevant information (e.g., identity and background) rather than subtle forgery traces.}

\item{We innovatively propose a simple yet effective intervention method SeLop from the perspective of causal representation learning. By applying an orthogonal low-rank intervention to the original CLIP representation space, we effectively eliminate the influence of spurious correlations unrelated to forgery traces, forcing the model to rely on true causal features for decision-making.}

\item{Extensive experiments on several standard benchmarks demonstrate the superior generalization and robustness of our method. In particular, with only 0.39M trainable parameters, our method achieves state-of-the-art performance.}
\end{enumerate}

The rest of this paper is organized as follows. Section II introduces the related work. In Section III, the proposed method is introduced in detail. Extensive experiment results and analysis are given in Section IV. Section V is the conclusion.

\section{Related Work}
In this section, we introduce traditional face forgery detection methods, face forgery detection methods based on pretrained visual foundation models, and methods that incorporate Causal Representation Learning (CRL).

\subsection{Traditional face forgery detection}
The rapid advancement of artificial intelligence technologies has spurred diversification in facial forgery techniques. Presently, improving generalizability of detection models remains a challenging issue in face forgery detection tasks. Previous work has focused on meticulously designing deep neural networks to learn ``specific and fake patterns", including biological signals, frequency anomalies, boundary artifacts and so on. LipForensics \cite{lips} trains a temporal network using fixed mouth embeddings from both real and fake data to detect fake videos based on mouth movements without overfitting to low-level, tamper-specific artifacts. DFGaze \cite{gaze} uncovers subtle forgery cues by analyzing differences in facial gaze angles between real and fake videos. F3Net \cite{f3} proposes a two-stream collaborative learning framework that leverages two different but complementary frequency-aware clues to deeply mine forgery patterns. F2trans \cite{f2trans} combines center difference attention and wavelet transform to extract high-frequency fine-grained features, which improves detection performance effectively. Tall \cite{tall} performs fixed-position masking on consecutive frames within each frame and rearranges them into a predefined layout as thumbnails, effectively uncovering spatiotemporal clues in forged videos. FTCN-TTN \cite{FTCNTTN} proposes a novel two-stage framework of temporal convolution network and temporal transformer network, aiming to explore the long-term temporal coherence.

However, due to the limited forgery patterns in training datasets, models often overfit and fail to generalize to unknown forgery techniques. To mitigate overfitting, \cite{bi,sbi} proposed a pseudo-fake face generation strategy. Training models on custom datasets improved generalization to some extent. However, expecting models to learn all forgery patterns on custom data remains impractical.

\subsection{Face forgery detection based on pretrained visual foundation models}
With the advent of large language models (LLMs), pre-trained visual foundation models (e.g., CLIP \cite{clip}) on massive data have demonstrated strong generalization across downstream tasks. Recent research has explored adapting CLIP for face forgery detection. UDD \cite{udd} designed a shuffle branch and a mix branch to break position and content biases of the CLIP. GM-DF \cite{GMDF} proposes a domain-aware hybrid expert modelling approach within a meta-learning framework, achieving generalized multi-scenario detection. FCG \cite{fcg} has thoroughly explored spatial and temporal clues in deepfake videos. MFCLIP \cite{mfclip} employs textual prompts to guide the learning of forgery traces. FFD-STA \cite{FFD-STA} proposed an innovative video-level blending approach to extract spatial and temporal features. RepDFD \cite{gaints} injects perturbations into input images to reprogram pre-trained VLM models. Forensic-Adapter \cite{FA} introduces an adapter specifically to learn blending boundaries, achieving excellent generalization through interaction between CLIP and the adapter. Effort \cite{orthogonal} employs Singular Value Decomposition (SVD) to decouple the feature space of attention, preserving pre-trained knowledge while learning forgery patterns. These methods explore various ways to adjust CLIP for face forgery detection and achieve good results.

\subsection{Causal Representation Learning}
In recent years, Causal Representation Learning (CRL) \cite{CRL} has attracted significant attention in the field of computer vision. Mitigating spurious correlations through causal intervention has proven a promising approach to improving cross-domain generalization \cite{causal}. Recently, some researchers have introduced the idea of CRL into face forgery detection. CADDM \cite{caddm} discovered and quantified the negative impact of identity information on deepfake detection, and ultimately improved the model's generalization ability by mitigating spurious correlation caused by identity. ID3 \cite{causalpaper1} jointly trained a purified invariant predictor and learned an aligned invariant representation, which forced the model to rely on domain-invariant features for determination, rather than being affected by forgery-irrelevant factors. UDD \cite{udd} identified position and content biases in CLIP \cite{clip}, designing shuffling and mixing branches to break inherent biases and achieve great generalization. CausalCLIP \cite{causalclip-causalpaper2} learned stable causal features robust to distribution shifts by modeling the generation process with a structural causal model and enforcing statistical independence. The success of these methods suggest that mitigating spurious correlations does indeed help improve generalization.

However, as spurious correlations arise from unobserved confounding factors, accurately identifying and intervening in all spurious correlations is impractical. Therefore, this paper proposes to intervene in the visual representations of pre-trained foundation models (e.g., CLIP) to mitigate spurious correlations.
\begin{figure}[t]
  \centering
  \includegraphics[width=\columnwidth]{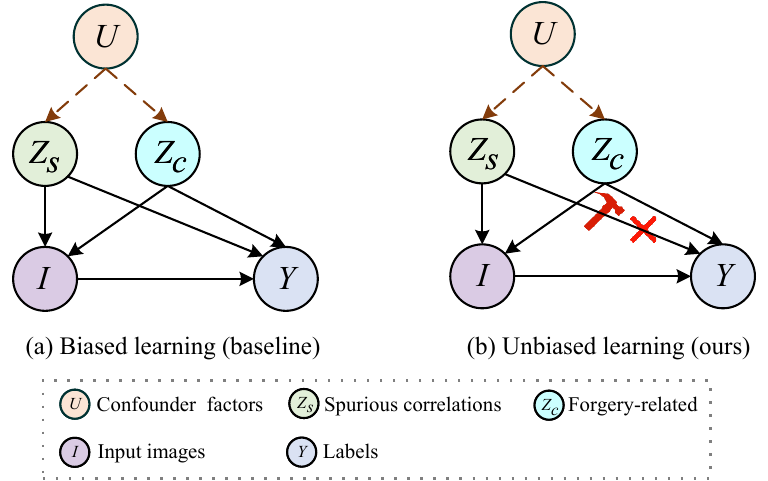}
  \caption{Causal analysis. Baseline fits to spurious correlations, relying on statistical characteristics of labels and forgery-irrelevant information for classification decisions. There is a backdoor path  $U\!\rightarrow\!Z_s\!\rightarrow\!Y$. However, our method cuts off the shortcut after successful intervention, forcing the model to truly rely on forgery traces for classification.}
  \label{fig:method1}
\end{figure}

\begin{figure*}[t]
  \centering
  \includegraphics[width=\textwidth]{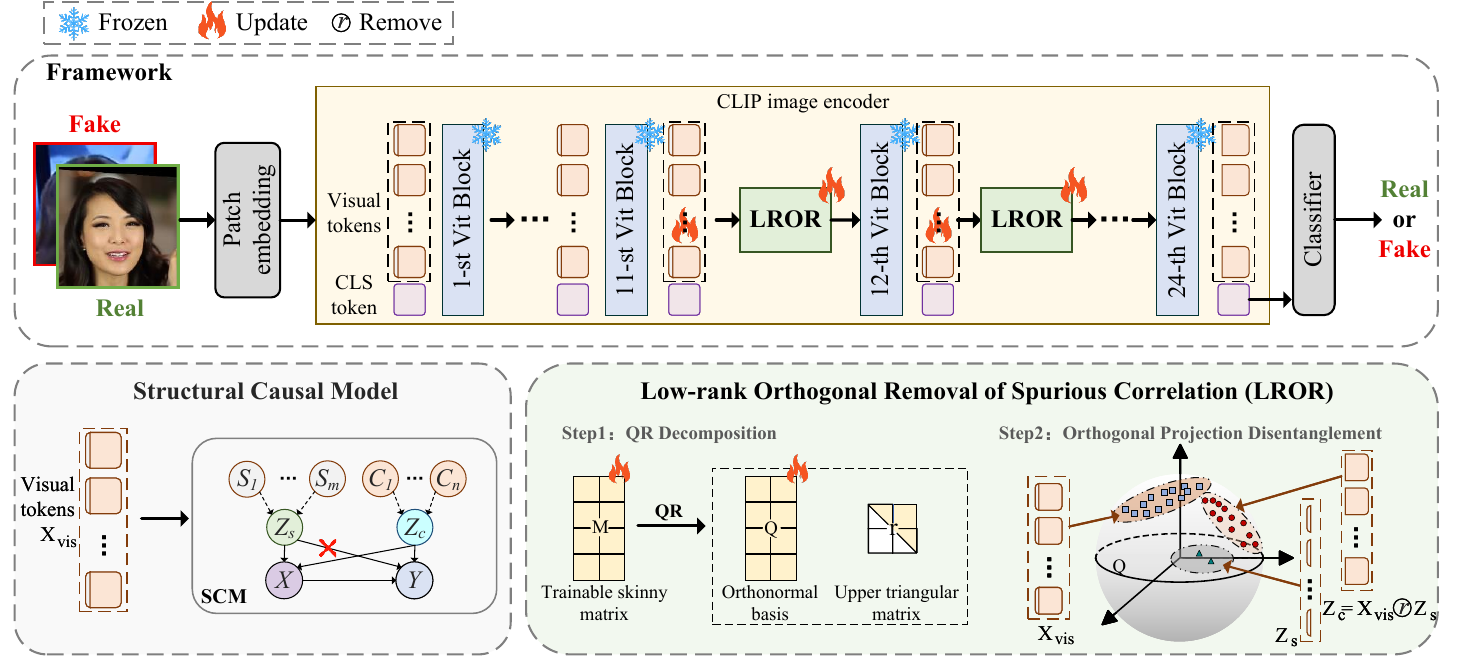}
  \caption{\textbf{Pipeline of the proposed method SeLop.} Based on the Structural Causal Model, we propose Low-rank Orthogonal Removal of Spurious Correlation (LROR) to mitigate influence of forgery-irrelevant factors $Z_s$, which aims to break the backdoor path from spurious correlations to labels. See text for details.}
  \label{fig:main}
\end{figure*}

\section{Methodology}
In this section, we first introduce our modeling approach in detail through III.A (Problem Formulation). Then, we introduce our specific method (Low-rank Orthogonal Removal of Spurious Correlation). Finally, we explain why our method effectively mitigates the impact of spurious correlations.

\subsection{Problem Formulation from Causal Representation Learning}
Face forgery detection is typically modeled as a binary classification problem, wherein the task involves determining whether a given face image is authentic or forged. Given a face image $I$ and a label $Y\in\{0,1\}$ (0 = real, 1 = fake), our goal is to learn a classifier that minimizes the distance between the predicted label and the ground-truth label, which can be formalized as:
\begin{equation}
\min_{\phi,\,f}\ \mathbb{E}\big[\ell\big(f(\phi(I)),\,Y\big)\big],
\label{eq:erm}
\end{equation}
where $\phi(I)$ denotes visual representations of face image $I$, $f$ denotes a classifier, and $\ell$ denotes the loss function in the training process. The expectation $\mathbb{E}$ is taken over the data distribution.

In this paper, we formulate the problem of face forgery detection from a novel perspective of causal representation learning. We utilize Structural Causal Model (SCM) \cite{causality} to define the relevant variables and factors, as shown in Fig. \ref{fig:method1}. The data-generating process can be described by the structural equations:
\begin{equation}\label{eq:scm}
\begin{split}
Z_s := g_s(U,\varepsilon_s),\quad Z_c := g_c(U,\varepsilon_c),\\
R := h(Z_s,Z_c,\varepsilon_r),\quad Y := r(Z_c,\varepsilon_y),
\end{split}
\end{equation}
where $\varepsilon_{\cdot}$ are exogenous noises. $g_s, g_c, h, r$ are structural functions in the SCM. For brevity, the encoder $h(\cdot)$ mixes $Z_s$ and $Z_c$ into the representation $R$ (the image formation and the feature extractor are folded into $h$). Finally, $r(Z_c,\varepsilon_y)$ defines the task orientation where the ground-truth label depends on forgery cues.

Let $U$ denote unobserved latent \emph{confounders}, $Z_s$ denote \emph{spurious correlation} factors (e.g., identity, background), and $Z_c$ denote \emph{causal} factors induced by manipulations (e.g., forgery traces). In this SCM, the real causal path $U\!\rightarrow\!Z_c\!\rightarrow\!Y$ carries forgery-related causal factors, whereas $U\!\rightarrow\!Z_s\!\rightarrow\!Y$ constitutes a backdoor path that induces spurious correlations in training process.

Driven by the \textit{low-rank spurious bias} phenomenon observed in Section I, we recognize that the variations of $Z_s$ dominate the principal components of the visual feature space, severely suppressing the subtle causal factors $Z_c$.
However, under ideal circumstance, the classifier should rely solely on forgery-related features to determine authenticity,
and be invariant to interference of forgery-irrelevant factors $Z_s$. Since $Z_s$ and $Z_c$ are unobserved during training, we impose these desiderata at the representation level. Given an input face image $I$, we write its representation as $R=\phi(I)= h(Z_s,Z_c,\varepsilon)$. 
Intuitively, the learned representation should retain enough information about the forgery-related factor $Z_c$ that determines $Y$, while being insensitive to spurious factors $Z_s$ or discard them. In the next section, we describe a transformation on $\phi(I)$ that achieves these desiderata.

\subsection{Low-rank Orthogonal Removal of Spurious Correlation}
To address \textit{low-rank spurious bias} phenomenon and effectively block the backdoor path $U \rightarrow Z_s \rightarrow Y$, we propose an orthogonal low-rank projection to intervene in the visual space of CLIP. Our goal is to enable the model to adaptively and correctly remove the influence of spurious correlation features $Z_s$ in the visual representation through training, and utilize the forgery-related features $Z_c$ to make classification decisions.

In order to retain CLIP's rich pre-training knowledge to the maximum extent, we only apply the projection intervention in the middle and deep layers of the CLIP to estimate a spurious correlation subspace. (see Fig. \ref{fig:main}). 

Specifically, let $X\in\mathbb{R}^{B\times(1+N_p)\times D}$ denote CLIP tokens in every transformer layer. Note that CLIP tokens $X$ comprise visual tokens $X_{vis}\in\mathbb{R}^{B\times N_p\times D}$ and [CLS] token $X_{[CLS]}\in\mathbb{R}^{B\times1\times D}$, denoted as $X=[X_{vis}, X_{[CLS]}]$. $B$ denote batchsize of the training process, $N_p$ denote number of visual tokens and $D$ denote hidden dimension of CLIP tokens. Consider all visual tokens as a visual space $X_{vis}\in\mathbb{R}^{N_p\times D}$, which is equivalent to the visual representation $\phi(I)$ that we discussed earlier. Intervention at each layer is implemented with a trainable skinny matrix $\mathbf{M}\in\mathbb{R}^{D\times r}$. Firstly, we perform $QR$ decomposition on $\mathbf{M}$ to obtain an orthonormal basis $\mathbf{Q}\in\mathbb{R}^{D\times r}, r \ll D$: 
\begin{equation}
\mathbf{Q},\mathbf{r}=QR(\mathbf{M})
\end{equation}
where $\mathbf{Q}^\top\mathbf{Q}=I_r$, $\mathbf{Q}$ is a column-orthogonal matrix and $\mathbf{r}$ is an upper triangular matrix. 
According to our PCA analysis, forgery-irrelevant signals exhibit a low-rank manifold distribution. Therefore, we can explicitly estimate the representation of spurious correlation $Z_s$ by decomposing the visual tokens into a low-rank compression subspace:
\begin{equation}
Z_{\text{s}}=X_{vis}\mathbf{Q}\mathbf{Q}^\top
\end{equation}

By actively projecting the features into this low-rank subspace through training, the redundant, macroscopic semantics (e.g., identity and background) are isolated. Consequently, the causal factors $Z_c$ associated with authentic forgery artifacts are disentangled and preserved within the orthogonal complement space:
\begin{equation}
Z_{\text{c}}=X_{vis}-Z_{\text{s}}
\label{eq:token-proj}
\end{equation}

By intervening in the visual representation, we suppress the effects of spurious correlation. 
Finally, we concat $Z_{\text{c}}$ and $X_{[CLS]}$ to form a new token stream $[Z_{\text{c}}, X_{[CLS]}]$ and feed it to the next transformer layer. See Algorithm 1 for the specific pseudocode. During training, we only optimize $\mathbf{Q}$, and the final fully-connected layer while keeping the parameters of all transformer block fixed. This implementation ensures the preservation of pre-trained knowledge of CLIP while reducing the introduction of extraneous noise. Formally, our model learns $\mathbf{Q}$ to push $X_{vis}\mathbf{Q}\mathbf{Q}^\top$ to align with spurious variation from $Z_s$ and removes it, preserving causal evidence from $Z_c$ ($U\!\rightarrow\!Z_c\!\rightarrow\!Y$)
and blocking the backdoor $U\!\rightarrow\!Z_s\!\rightarrow\!Y$. 

\textbf{Objective.} After intervening in the visual space of CLIP, our goal can be formulated as:
\begin{equation}
\begin{aligned}
\min_{\phi,\,f,\,\mathbf{Q}}\;&
\mathbb{E}_{(X,Y)}\!\Big[
  \ell\!\big(f\big(\,\phi(I)\,P_{\perp}\big),\, Y\big)
\Big] \\[2pt]
\text{s.t. }\;&
P_{\perp}= I-\mathbf{Q}\mathbf{Q}^{\top},
\mathbf{Q}^{\top}\mathbf{Q}=I_r
\end{aligned}
\label{eq:proj-erm-row}
\end{equation}

In the training phase, the main objective function is a Cross-entropy loss applied on $X_{[CLS]}$ in the last layer of CLIP.

\begin{algorithm}[ht]
\caption{Low-rank Orthogonal Removal of Spurious Correlation}
\begin{algorithmic}[1]
\Require Original token stream $X=[X_{vis}, X_{[CLS]}] \in \mathbb{R}^{B \times (1+N_p) \times D}$.
\Ensure Intervened token stream
\Statex \hphantom{\textbf{Output: }} $[Z_c, X_{[CLS]}] \in \mathbb{R}^{B \times (1+N_p) \times D}$.
\State Learnable basis $\mathbf{B} \in \mathbb{R}^{D \times r}$, $\mathbf{Q}, \mathbf{r} \leftarrow \mathrm{QR}(\mathbf{B})$ ;
\Statex \texttt{// QR decomposition}
\State $X_{\mathrm{CLS}} \leftarrow X[:, :1, :], X_{\mathrm{vis}} \leftarrow X[:, 1:, :]$ ;
\Statex \texttt{// Shape:\quad} $[B, 1, D]$ \texttt{and} $[B, N_p, D]$
\State Define low-rank projector $P = \mathbf{Q}\mathbf{Q}^\top, P_{\perp} = I - P$
\State $Z_s \leftarrow (X_{\mathrm{vis}} \mathbf{Q}) \mathbf{Q}^\top$ ; \hfill \texttt{// }$= X_{\mathrm{vis}} P$
\State $Z_c \leftarrow X_{\mathrm{vis}} - Z_s$ ; \hfill \texttt{// }$= X_{\mathrm{vis}} P_{\perp}$
\State $[Z_c, X_{[CLS]}] \leftarrow \mathrm{concat}(\mathrm{CLS}, Z_c)$
\end{algorithmic}
\end{algorithm}

\subsection{Why our method works}
We model the spurious correlation subspace by a column-orthonormal basis \(Q \in \mathbb{R}^{D \times r}\) with \(r \ll D\), obtained by orthogonalizing a slim learnable matrix \(B\) via QR decomposition to get \(Q\), which induces the projector \(P = Q Q^\top\) and its orthogonal complement \(P_\perp = I - P\). 

At both training and inference, we remove spurious correlations by applying \(P_\perp\) to token features. Under the row-wise convention for features \(X \in \mathbb{R}^{N \times D}\), we compute \(Z_c = X P_\perp = X (I- P)\), thus preserving forgery-related signal while suppressing spurious directions. The orthogonal basis \(Q\) is learned end-to-end purely from the cross-entropy loss: if \(Q\) captures shortcuts $Z_s$, projecting them out improves generalization; if it mistakenly captures causal factors $Z_c$, the loss penalizes it and drives \(Q\) away. This yields a precise mechanism to excise spurious correlation subspaces without any explicit supervision of \(Z_s\) or \(Z_c\).

\section{Experiments}
In this section, we first introduce the datasets used in the subsequent experiments. Next, we describe the implementation
details of SeLop. We then discuss the experimental results of generalization evaluation, including cross-dataset evaluation, cross-manipulation method evaluation to demonstrate the effectiveness of the proposed method. Moreover, we discuss the robustness of SeLop under different perturbations. Additionally, we conduct ablation experiments to validate the effectiveness of LROR by qualitative and quantitative analysis. Finally, we conduact hyperparameter sensitivity analysis and complexity analysis of SeLop.

\subsection{Experimental Settings}
\noindent{\textbf{Datasets.}} To comprehensively and fairly evaluate the model's generalization ability, we follow the DeepFakeBench protocol \cite{deepfakebench} and utilize six publicly available datasets: FaceForensics++ (FF++)\cite{rossler2019faceforensics++}, Celeb-DF (CDF-v1 and CDF-v2)\cite{li2020celeb}, Deepfake Detection Challenge dataset (DFDC)\cite{dfdc}, DeepFake Detection Challenge Preview (DFDCP)\cite{dfdcp}, and Deepfakedetection (DFD)\cite{Dfd}.

\subsubsection{FaceForensics++} FaceForensics++ \cite{rossler2019faceforensics++} is a standard benchmark in face forgery detection. It includes 1,000 real videos and 4,000 fake videos, which consist of four types of face manipulation techniques: Deepfakes (DF) \cite{Deepfakes}, Face2Face (F2F) \cite{thies2016face2face}, FaceSwap (FS) \cite{faceswap}, and NeuralTextures
(NT) \cite{nt}. In our experiments, we use the c23 compression version of FF++ to train and evaluate all the methods.

\subsubsection{Celeb-DF} Celeb-DF-v1 \cite{li2020celeb} includes 408 real videos and 795 synthesized videos with reduced visual artifacts. Celeb-DF-v2 \cite{li2020celeb} is a larger dataset forged by several forgery techniques and is an extension of the Celeb-DF-v1 version. It consists of 590 real videos, featuring subjects of different ages, races, and along with 5639 corresponding DeepFake videos.

\subsubsection{Deepfake Detection Challenge dataset} DFDC \cite{dfdc} is a challenging large-scale video dataset, since its real videos are very close to real life, and the forgery traces are smaller and more subtle. It contains over 100,000 videos, so we only conduct cross-dataset evaluation on its test set. DFDCP \cite{dfdcp} is a dataset of 5,000 videos created by Facebook to advance deep forgery detection technology, which is the preview version of the Deepfake Detection Challenge (DFDC) dataset.

\subsubsection{DeepFakeDetection} DFD \cite{Dfd} consists of 363 real videos and 3071 synthetic videos, which is generated by Google and collected in FF++ now.

In this paper, we adopt four widely used and standard protocols for evaluation: \textbf{Protocol-1}: cross-dataset evaluation, \textbf{Protocol-2}: cross-manipulation evaluation on the latest DF40 and on FF++ (c23). \textbf{Protocol-3}: Intra-dataset evaluation. \textbf{Protocol-4}: detection in real-world scenarios. For \textbf{Protocol-1}, all methods are trained on FF++ c23 training set and evaluated on other deepfake datasets. For \textbf{Protocol-2}, on one hand, we evaluate the models on the latest deepfake dataset DF40 \cite{df40}, which includes four types of forgery methods: face swapping, face reenactment, whole face synthesis, and face editing. On the other hand, we train models on one manipulation type of FF++ and test them on three unknown forgery techniques. For \textbf{Protocol-3}, all methods are trained on Celeb-DF-v2 dataset and evaluated on it too. For \textbf{Protocol-4}, we evaluate our model on latest datasets, DDL \cite{ddl}. It offers more diverse forgery methods and facial scenes, serving as more challenging benchmarks for real-world scenarios.

\begin{table*}[t]
  \caption{Cross-dataset evaluation results (\textbf{Frame-level AUC}). The \textbf{best} results are indicated in bold and the \underline{second-best} results are underlined. ``$^{\dagger}$" indicates that we reproduce results with its original code.}
  \label{tab:frame-level}
  \centering
  \setlength{\tabcolsep}{2.5pt}
  \renewcommand{\arraystretch}{1.0}
  
  \resizebox{0.75\textwidth}{!}{
  \begin{tabular}{c|c|ccccc|c}
    \toprule
    Method & Venue & CDF-v1 & CDF-v2 & DFDC &
    DFDCP & DFD & Avg.\\
    \midrule
    Xception \cite{xception}   & ICCV’19 & 0.779 & 0.737 & 0.708 & 0.737 & 0.816 & 0.755\\
    F3Net \cite{f3}            & ECCV’20 & 0.777 & 0.735 & 0.702 & 0.735 & 0.798 & 0.749\\
    SPSL \cite{spsl}           & CVPR’21 & 0.815 & 0.765 & 0.704 & 0.741 & 0.812 & 0.767\\
    SRM \cite{srm}             & CVPR’21 & 0.793 & 0.755 & 0.700 & 0.741 & 0.812 & 0.760\\
    Recce \cite{recce}         & CVPR’22 & 0.768 & 0.732 & 0.713 & 0.734 & 0.812 & 0.752\\
    UCF \cite{ucf}             & ICCV’23 & 0.779 & 0.753 & 0.719 & 0.759 & 0.807 & 0.763\\
    ED \cite{ed}               & AAAI’24 & 0.818 & 0.864 & 0.721 & 0.851 &  -    &  - \\
    CFM \cite{cfm}             & TIFS’24 & -     & 0.828 & -     & 0.758  & 0.915 & - \\
    LSDA \cite{lsda}           & CVPR’24 & 0.867 & 0.830 & 0.736 & 0.815 & 0.880 & 0.826\\
    UDD \cite{udd}             & AAAI’25 & -     & 0.869 & 0.758 & 0.856 & 0.910 & -\\
    FIA-USA \cite{FIAUSA}      & arXiv’25 & 0.901 & 0.867 & -    & 0.818 & 0.821 & -\\
    GM-DF \cite{GMDF}          & ACM MM’25 & 0.893 & 0.892 & \underline{0.847} & 0.882 & 0.928 & 0.888\\
    Effort$^{\dagger}$ \cite{orthogonal}   & ICML’25 & \underline{0.915} & 0.898 & 0.825 & 0.860 & \underline{0.932} & 0.886\\
    Forensics-Adapter \cite{FA} & CVPR’25 & 0.914 & \underline{0.900} & 0.843 & \underline{0.890} & \textbf{0.933} & \underline{0.896}\\
    \midrule
    \rowcolor{blue!5}
     SeLop (Ours)    & -  & \textbf{0.924} & \textbf{0.901} & \textbf{0.853} & \textbf{0.905} & 0.926 & \textbf{0.902}\\
    \bottomrule
  \end{tabular}}
\end{table*}

\begin{table}[t]
  \caption{Cross-dataset evaluation results (\textbf{Video-level AUC}). All the results are taken from their original papers.}
  \label{tab:video-level}
  \centering
  
  \setlength{\tabcolsep}{3.0pt} 
  \renewcommand{\arraystretch}{1.0}
  
  \resizebox{0.48\textwidth}{!}{
  \begin{tabular}{c|c|ccc}
    \toprule
    Method & Venue & CDF-v2 & DFDC & DFDCP \\
    \midrule
    SBI \cite{sbi}             & CVPR’22 & 0.932 & 0.724 & 0.862 \\
    RealForensics \cite{realforensics}   & CVPR’22 & 0.869 & 0.759 & - \\
    AUNet \cite{aunet}           & CVPR’23 & 0.928 & 0.738 & 0.862 \\
    SeeABLE \cite{seeable}       & ICCV’23 & 0.873 & 0.759 & 0.863 \\
    CADDM \cite{caddm}           & CVPR’23 & 0.939 & 0.739 & - \\
    IID \cite{iid}             & CVPR’23 & 0.838 & - & 0.812 \\
    LAA-Net \cite{laa}         & CVPR’24 & \underline{0.954} & - & 0.869 \\
    TALL++ \cite{talljiajia}    & IJCV’24 & 0.920 & 0.785 & - \\
    MOE-FFD \cite{Moeffd}      & TDSC’25 & 0.913 & - & 0.850 \\
    Effort \cite{orthogonal}   & ICML’25 & \textbf{0.956} & \underline{0.843} & \underline{0.909} \\
    \midrule
    \rowcolor{blue!5}
     SeLop (Ours)    & - &
    0.948 & \textbf{0.877} & \textbf{0.929}\\
    \bottomrule
  \end{tabular}}
\end{table}

\begin{table}[t]
  \caption{Cross-dataset evaluation results (\textbf{Video-level AUC}). Comparing our method with other CLIP-based methods.}
  \label{tab:CLIP-based method}
  \centering
  
  \setlength{\tabcolsep}{3.0pt} 
  \renewcommand{\arraystretch}{1.0}
  \resizebox{0.48\textwidth}{!}{
  \begin{tabular}{c|c|ccc}
    \toprule
    Method & Venue & CDF-v2 & DFDC & DFDCP \\
    \midrule
    Vanilla CLIP\cite{clip}             & ICML’21 & 0.809 & 0.766 & 0.785 \\
    \midrule
    FFAA \cite{ffaa}                     & arXiv’24 & - & 0.740 & - \\
    MFCLIP \cite{mfclip}                 & TIFS’25 & 0.835 & \underline{0.861} & - \\
    UDD \cite{udd}             & AAAI’25 & 0.931 & 0.812 & 0.881 \\
    FCG \cite{fcg}                       & CVPR’25 & \textbf{0.950} & 0.818 & - \\
    FFD-STA \cite{FFD-STA}               & CVPR’25 & 0.947 & 0.843 & \underline{0.909} \\
  
    \midrule
    \rowcolor{blue!5}
     SeLop (Ours)    & - &
    \underline{0.948} & \textbf{0.877} & \textbf{0.929}\\
    \bottomrule
  \end{tabular}}
\end{table}

\noindent{\textbf{Evaluation Metrics.}} Following the most existing work \cite{lsda, udd, FA}, we utilize area under receiver operating characteristic curve (AUC) scores for generalization evaluation in experiments. In our paper, we report results for both frame-level AUC and video-level AUC. Frame-level AUC is calculated based on the predicted scores of face frames. In line with these methods, video-level AUC is computed by averaging the model’s outputs across frames of a video.

\noindent{\textbf{Implementation Details.}} Our method is implemented by Pytorch 1.12.0 with one Nvidia GTX 4090 GPU. We utilize CLIP ViT-L/14 \cite{clip} as the default vision foundation model. In subsequent experiments, we also verify the effect on other versions of CLIP. Following the configuration of DeepFakeBench \cite{deepfakebench}: For each video, 32 frames were extracted for training or testing, and RetinaFace \cite{retinaface} was used for face extraction. We use the fixed learning rate of 2e-4 and weight decay of 0.0005 for training our model on FF++ \cite{rossler2019faceforensics++} c23 training set. Adam optimiser is employed for optimization. We set the batch size to 32 for both training and testing. In addition, the rank for the low-rank spurious correlation subspace is set to 32 and the last 12 layers serve as the target layers for our intervention. To ensure comprehensive and fair comparisons, all our experiments were conducted under the DeepFakeBench framework \cite{deepfakebench}.

\subsection{Comparison with State-of-the-art Methods}
In this section, we conduct cross-dataset and intra-dataset experiments to evaluation our method and compare SeLop with other SOTA methods.

\noindent\textbf{Protocol-1: Cross-dataset Evaluation.} We report frame-level and video-level results. Furthermore, we compare our method with state-of-the-art CLIP fine-tuning based  methods to demonstrate the superiority of SeLop low-rank interventions.
\subsubsection{Frame-level Evaluation} Table \ref{tab:frame-level} shows the cross-dataset evaluation results in frame-level AUC. The results for ED \cite{ed}, CFM \cite{cfm}, and FIA-USA \cite{FIAUSA} are directly cited from the original paper and other results are cited from Forensics-Adapter \cite{FA} under DeepFakeBench \cite{deepfakebench}. Effort \cite{orthogonal} do not report frame-level results, so we reproduce them using its open source code. From the table, we can observe that our approach outperforms the current state-of-the-art methods Effort and Forensics-Adapter. On the widely regarded most challenging DFDC and DFDCP datasets, our method surpasses Forensics-Adapter by 1\% (from 0.843 to 0.853) and 1.5\% (from 0.890 to 0.905) AUC respectively. These results demonstrate the excellent generalization performance of our method.

\begin{table*}[t]
  \caption{Cross-manipulation evaluations (Video-level AUC). We evaluate all methods on the latest DF40\cite{df40} dataset. The best results are indicated in bold and the second-best results are underlined.}
  \label{tab:cross-manipulation}
  \centering
  
  \setlength{\tabcolsep}{3.0pt} 
  \renewcommand{\arraystretch}{1.1}
  \resizebox{0.75\textwidth}{!}{
  \begin{tabular}{c|cccccccc|c}
    \toprule
    Method  &
    UniFace & BleFace & MobSwap & e4s & FaceDan & FSGAN & InSwap & SimSwap & Avg.\\
    \midrule
    F3Net \cite{f3}                 & 0.809 & 0.808 & 0.867 & 0.494 & 0.717 & 0.845 & 0.757 & 0.674 & 0.746\\
    SPSL \cite{spsl}                & 0.747 & 0.748 & 0.885 & 0.514 & 0.666 & 0.812 & 0.643 & 0.665 & 0.710\\
    SRM \cite{srm}                  & 0.749 & 0.704 & 0.779 & 0.704 & 0.659 & 0.772 & 0.793 & 0.694 & 0.732\\
    SBI \cite{sbi}                  & 0.724 & \underline{0.891} & 0.952 & 0.750 & 0.594 & 0.803 & 0.712 & 0.701 & 0.766\\
    IID \cite{iid}                  & 0.839 & 0.789 & 0.888 & 0.766 & \underline{0.844} & 0.927 & 0.789 & 0.644 & 0.811\\
    LSDA \cite{lsda}                & 0.872 & 0.875 & 0.930 & 0.694 & 0.721 & 0.939 & 0.855 & 0.793 & 0.835\\
    CDFA \cite{cdfa}                & 0.762 & 0.756 & 0.823 & 0.631 & 0.803 & \underline{0.942} & 0.772 & 0.757 & 0.781\\
    ProDet \cite{ProDet}            & 0.908 & \textbf{0.929} & \textbf{0.975} & 0.771 & 0.747 & 0.928 & 0.837 & 0.844 & \underline{0.867}\\
    FIA-USA \cite{FIAUSA}           & \underline{0.918} & -     & -     & \underline{0.875} & 0.830 & 0.863 & \underline{0.874} & \textbf{0.910} & -    \\
    MAP-Mamba \cite{mapmamba}       & 0.815 & 0.896 & \underline{0.959} & 0.779 & 0.826 & - & 0.855 & 0.813 & - \\
    \midrule
    \rowcolor{blue!5}
    SeLop (Ours)    &  \textbf{0.925} & 0.866 & 0.957 & \textbf{0.940} & \textbf{0.880} & \textbf{0.952} & \textbf{0.897} & \underline{0.856} & \textbf{0.909}\\
    \bottomrule
  \end{tabular}}
\end{table*}

\subsubsection{Video-level Evaluation} Table \ref{tab:video-level} shows the cross-dataset evaluation results in video-level AUC. We compare 10 state-of-the-art methods, with all results directly cited from their original papers. In video-level comparisons, our approach outperform the state-of-the-art Effort by 3.4\% (from 0.843 to 0.877) and 2\% (from 0.909 to 0.929) AUC on the DFDC and DFDCP datasets respectively, providing compelling evidence of its superior generalization.

\subsubsection{Compared with CLIP-based Methods} Table \ref{tab:CLIP-based method} presents a comparison of our method with other CLIP-based methods. It can be observed that our method outperforms the second-best approach by 1.6\% (from 0.861 to 0.877) and 2\% (from 0.909 to 0.929) on the DFDC and DFDCP datasets respectively. It is noteworthy that our approach significantly outperforms UDD, which specifically distinguishes and addresses one particular type of spurious correlation. The results further demonstrate the effectiveness of our approach.

\begin{table}[t]
  \caption{Cross-manipulation evaluation results (Video-level AUC). Cross Avg. represents the average results on three cross-manipulation evaluation trials.}
  \label{tab:cross-FF++}
  \centering
  
  \resizebox{0.48\textwidth}{!}{
  \begin{tabular}{c|c|ccccc}
  \toprule
  Methods & Train & DF & F2F & FS & NT & Cross Avg. \\
  \midrule
  EN-B4 \cite{efficientnet} & \multirow{3}{*}{DF} & 0.999 & 0.683 & 0.452 & 0.657 & 0.597 \\
  RECCE \cite{recce} & & 0.999 & 0.698 & 0.547 & 0.772 & 0.672 \\
  CFM \cite{cfm} & & 0.999 & 0.776 & 0.549 & 0.750 & 0.692 \\
  SeLop & & \cellcolor{blue!5}0.999 & \cellcolor{blue!5}0.753 &  \cellcolor{blue!5}0.980 & \cellcolor{blue!5}0.637 & \cellcolor{blue!5}\textbf{0.790}\\
  \midrule
  EN-B4 \cite{efficientnet} & \multirow{3}{*}{F2F} & 0.818 & 0.991 & 0.583 & 0.665 & 0.688 \\
  RECCE \cite{recce} & & 0.716 & 0.992 & 0.500 & 0.723 & 0.646 \\
  CFM \cite{cfm} & & 0.819 & 0.992 & 0.601 & 0.708 & 0.709 \\
  SeLop & & \cellcolor{blue!5}0.948 & \cellcolor{blue!5}0.997 & \cellcolor{blue!5}0.907 & \cellcolor{blue!5}0.630 & \cellcolor{blue!5}\textbf{0.828}\\
  \midrule
  EN-B4 \cite{efficientnet} & \multirow{3}{*}{FS} & 0.682 & 0.669 & 0.996 & 0.512 & 0.621 \\
  RECCE \cite{recce} & & 0.631 & 0.662 & 0.997 & 0.581 & 0.624 \\
  CFM \cite{cfm} & & 0.729 & 0.714 & 0.999 & 0.517 & 0.653 \\
  SeLop & & \cellcolor{blue!5}0.993 & \cellcolor{blue!5}0.839 & \cellcolor{blue!5}0.999 & \cellcolor{blue!5}0.565 & \cellcolor{blue!5}\textbf{0.799}\\
  \midrule
  EN-B4 \cite{efficientnet} & \multirow{3}{*}{NT} & 0.821 & 0.750 & 0.493 & 0.991 & 0.688 \\
  RECCE \cite{recce} & & 0.724 & 0.647 & 0.516 & 0.996 & 0.629 \\
  CFM \cite{cfm} & & 0.883 & 0.768 & 0.526 & 0.992 & 0.726 \\
  SeLop & & \cellcolor{blue!5}0.895 & \cellcolor{blue!5}0.820 & \cellcolor{blue!5}0.711 & \cellcolor{blue!5}0.979 & \cellcolor{blue!5}\textbf{0.809}\\
  \bottomrule
  \end{tabular}}
\end{table}

\begin{table}[t]
  \caption{Intra-dataset comparison in terms of ACC and AUC. Comparison with State-of-the-art methods on Celeb-DF-v2 dataset. The best results are indicated in bold and the second-best results are underlined.}
  \label{tab:intra-dataset-CDF2}
  \centering
  
  \setlength{\tabcolsep}{4.0pt} 
  \renewcommand{\arraystretch}{1.3}
  
  \begin{tabular}{c|cc}
  \toprule
  \multirow{2}{*}{Method} & \multicolumn{2}{c}{CelebDF-v2} \\
  \cline{2-3}
  & ACC  & AUC  \\
  \midrule
    F3-Net \cite{f3} & \underline{0.960} & 0.989 \\
    Two-branch \cite{two-branch} & 0.938 & 0.959 \\
    TD3DCNN \cite{TD3DCNN} & 0.811 & 0.888 \\
    SRM \cite{srm} & 0.883 & 0.967 \\
    PEL \cite{PEL} & \textbf{0.968} & 0.982 \\
    STDT \cite{STDT} & 0.917 & 0.972 \\
    CORE  \cite{core} & 0.854 & 0.992  \\
    PCC \cite{PCC} & - & 0.990 \\
    LPS \cite{LPSssd} & - & 0.968 \\
    SPSL \cite{spsl} & 0.916 & \underline{0.996}  \\
    MADD \cite{MADD} & 0.946 & 0.980 \\
  \midrule
  \rowcolor{blue!5}
  \textbf{SeLop} & \textbf{0.968} & \textbf{0.999} \\
  \bottomrule
  \end{tabular}
\end{table}

\begin{table}[t]
  \caption{Face Forgery Detection in real-world scenes on the latest DDL \cite{ddl} dataset. We compare SeLop with eight SOTA methods.}
  \label{tab:DDL}
  \centering
  
  \setlength{\tabcolsep}{4.0pt} 
  \renewcommand{\arraystretch}{1.3}
  
  \begin{tabular}{c|cc}
  \toprule
  \multirow{2}{*}{Method} & \multicolumn{2}{c}{DDL} \\
  \cline{2-3}
  & ACC  & AUC  \\
  \midrule
    Xception \cite{xception} & 0.435 & 0.538 \\
    F3-Net \cite{f3} & 0.453 & 0.523 \\
    RFM \cite{rfm} & 0.523 & 0.573 \\
    SRM \cite{srm} & 0.560 & 0.503 \\
    CORE  \cite{core} & 0.443 & 0.535  \\
    SPSL \cite{spsl} & 0.607 & 0.613  \\
    MAP-Mamba \cite{mapmamba} & 0.672 & 0.712  \\
    ForensicAdapter \cite{FA} & \underline{0.779} & \underline{0.835} \\
  \midrule
  \rowcolor{blue!5}
  \textbf{SeLop} & \textbf{0.855} & \textbf{0.933} \\
  \bottomrule
  \end{tabular}
\end{table}

\begin{figure*}[t]
  \centering
  \includegraphics[width=\textwidth]{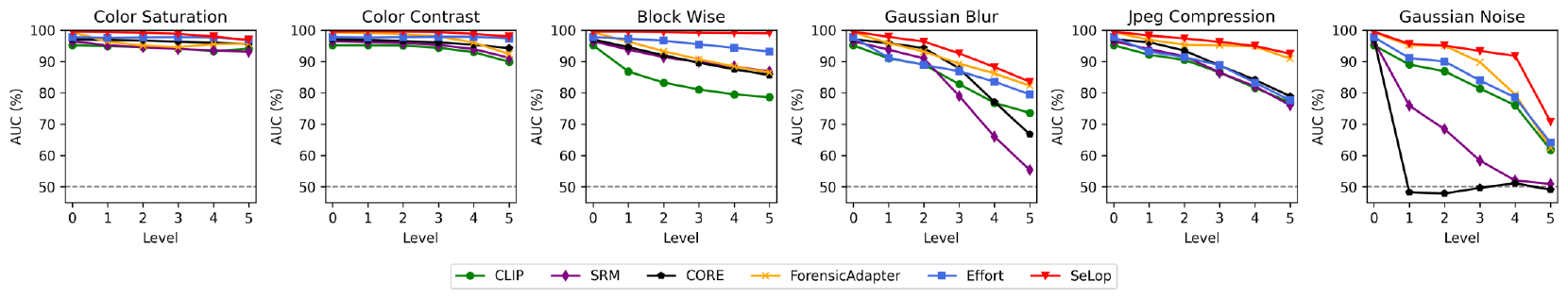}
  \caption{Robustness Analysis. Our method is compared with CLIP\cite{clip}, SRM\cite{srm}, CORE\cite{core}, ForensicAdapter \cite{FA}, Effort\cite{orthogonal} across five levels of six particular types of perturbations in \textbf{video-level AUC}.}
  \label{fig:robustness}
\end{figure*}


\noindent\textbf{Protocol-2: Cross-manipulation Evaluation} 

Table \ref{tab:cross-manipulation} shows cross-manipulation evaluation results. All models in the table were trained on the FF++ dataset and then tested on the latest DF40 \cite{df40} dataset. Note that DF40 includes multiple manipulation techniques (e.g., face swapping, face reenactment). It can be observed that our method achieves excellent detection performance when confronted with different manipulation techniques. In particular, our method outperforms the second-best method by 6.5\% (from 0.875 to 0.940) and 3.6\% (from 0.844 to 0.880) on the e4s \cite{e4s} and FaceDancer \cite{facedancer} datasets, respectively.

In addition, we conduct cross-manipulation experiments on FF++ (C23) to evaluate the detector’s generalization to unknown manipulation methods. We train the model on one forgery type and evaluate it on all four types (DF, F2F, FS, and NT). As shown in Table \ref{tab:cross-FF++}, when the models are trained on the FS type, SeLop outperforms CFM in cross-manipulation evaluations with a near 15\% Cross Avg. AUC improvement. This strongly demonstrates that our method does not overfit to any particular forgery pattern, but instead learns a general representation.

\noindent\textbf{Protocol-3: Intra-dataset Evaluation} 

To further demonstrate the effectiveness of our method on the intra-dataset, we compare it with other state-of-the-art methods on Celeb-DF-v2 \cite{li2020celeb}. As shown in Table \ref{tab:intra-dataset-CDF2}, we can see that SeLop outperforms all methods and achieves an impressive AUC performance of 0.999. This demonstrates SeLop's powerful ability to learn what is ``fake".

\noindent\textbf{Protocol-4: Detection in real-world scenarios} 

It is an important protocol in face forgery detection to evaluate models on benchmarks oriented towards real-world scenarios. As shown in Table \ref{tab:DDL}, it can be observed that our method outperforms all methods in terms of ACC and AUC. Surprisingly, SeLop achieves an image-level AUC of 0.933 on DDL dataset \cite{ddl}.

\begin{table*}[t]
\caption{Ablation study on visual space intervention. Evaluation metrics are the frame-level AUC, AP, and EER, respectively.}
\label{tab:subspace change}
\centering
\setlength{\tabcolsep}{8pt}
\renewcommand{\arraystretch}{1}

\resizebox{\textwidth}{!}{%
\begin{tabular}{c|c|c| ccc| ccc| ccc| ccc| ccc}
\toprule

\multirow{2}{*}{ID}&\multirow{2}{*}{SP} & \multirow{2}{*}{CA}
& \multicolumn{3}{c|}{CDF\mbox{-}v1}
& \multicolumn{3}{c|}{CDF\mbox{-}v2}
& \multicolumn{3}{c|}{DFDC}
& \multicolumn{3}{c|}{DFDCP}
& \multicolumn{3}{c}{DFD} \\
\cmidrule(lr){4-6}\cmidrule(lr){7-9}\cmidrule(lr){10-12}\cmidrule(lr){13-15}\cmidrule(lr){16-18}

& & &
AUC $\uparrow$ & AP $\uparrow$ & EER $\downarrow$ &
AUC $\uparrow$ & AP $\uparrow$ & EER $\downarrow$ &
AUC $\uparrow$ & AP $\uparrow$ & EER $\downarrow$ &
AUC $\uparrow$ & AP $\uparrow$ & EER $\downarrow$ &
AUC $\uparrow$ & AP $\uparrow$ & EER $\downarrow$ \\
\midrule

1&$\checkmark$ & $\checkmark$
& 0.781 & 0.859 & 0.299
& 0.763 & 0.862 & 0.309
& 0.742 & 0.770 & 0.326
& 0.761 & 0.869 & 0.316
& 0.828 & 0.978 & 0.248 \\
2&$\checkmark$ & $\times$
& 0.543 & 0.645 & 0.455
& 0.544 & 0.699 & 0.475
& 0.661 & 0.690 & 0.385
& 0.611 & 0.720 & 0.416
& 0.619 & 0.928 & 0.417 \\
3&$\times$ & $\checkmark$
& \textbf{0.924} & \textbf{0.954} & \textbf{0.160}
& \textbf{0.901} & \textbf{0.947} & \textbf{0.189}
& \textbf{0.853} & \textbf{0.883} & \textbf{0.233}
& \textbf{0.905} & \textbf{0.951} & \textbf{0.173}
& \textbf{0.926} & \textbf{0.991} & \textbf{0.149}\\

\bottomrule
\end{tabular}
}
\end{table*}

\noindent\textbf{Robustness Evaluation.} Figure \ref{fig:robustness} shows robustness evaluation results. Similar to the settings in \cite{udd}, we conduct robustness evaluation experiments using the FF++ test set. We employ the same configuration as in \cite{robustness}, including six different types of perturbations across five severity levels: color saturation, color contrast, block-wise, gaussian blur, JPEG compression, and gaussian noise. We compare our method with vanilla CLIP \cite{clip}, SRM \cite{srm}, and CORE \cite{core}. It is evident that our method exhibits greater robustness. These methods are highly sensitive to perturbation-related information, being significantly susceptible to them. However, our method can truly focus on forgery-related features after eliminating spurious correlations, and is less susceptible to irrelevant noise.

\subsection{Ablation Study}
\begin{figure}[t]
  \centering
  \includegraphics[width=\columnwidth]{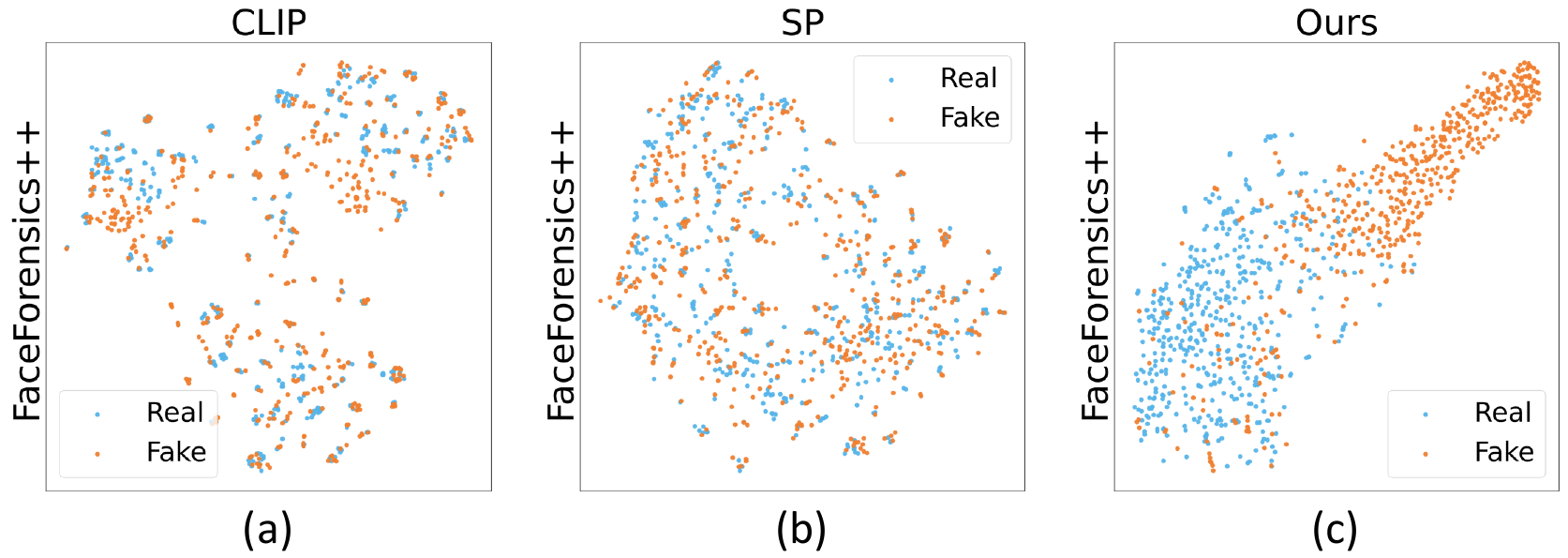}
  \caption{T-SNE Visualizations. The feature distribution before and after intervention in visual space of CLIP\cite{clip}: Vanilla CLIP (left, before intervention), spurious correlation subspace (mid), our method (right, after intervention).}
  \label{fig:tsne}
\end{figure}

\noindent \textbf{Effectiveness of LROR.} To verify that our method can project spurious correlations into a low-rank subspace and remove them from the original representation, we conduct a counterfactual validation experiment. Quantitative and qualitative experiments are shown in Table \ref{tab:subspace change} and Figure \ref{fig:tsne}. We replace the forward visual space with two complementary forms for comparative validation. The first is the non-causal factor space $\mathbf{X_{vis}}\mathbf{Q}\mathbf{Q}^\top$ (denoted by SP, representing the subspace of spurious correlations). We achieve this setting by fixing $\mathbf{Q}$ obtained from trained SeLop, replacing only the forward flow with $\mathbf{X_{vis}}\mathbf{Q}\mathbf{Q}^\top$, and retraining the linear classification head. The second is the causal factor space $\mathbf{X_{vis}}(\mathbf{I}-\mathbf{Q}\mathbf{Q}^\top)$ (denoted by CA, representing the visual space after our method's intervention). ID1 in Table \ref{tab:subspace change} is the Vanilla CLIP, without subspace replacement or intervention, where forgery-related knowledge is intertwined with irrelevant factors to forgery. As can be seen in Figure \ref{fig:tsne} (a), the distributions of real and forged samples overlap significantly, making it difficult to form a robust decision boundary. When the visual space consists solely of non-causal factors (corresponding to ID2 and Figure \ref{fig:tsne} (b)), the AUC performance across several datasets hovers around 0.5, with lower AP and higher error rate. This near-random performance suggests that this subspace primarily contains non-causal features, making it difficult for the model to accurately distinguish real from fake faces based on this information. However, after intervention with our method (corresponding to ID3 and Figure \ref{fig:tsne} (c)), the AUC and AP on each benchmark significantly improve, the EER significantly decreases, and T-SNE visualizations \cite{tsne} demonstrate clear between-class separation. This demonstrates that our method effectively removes the low-rank spurious correlation subspace induced by confounding factors at the representation level and focuses the spurious correlation factors on orthogonal complements, resulting in consistent improvements across datasets.

\begin{table*}[t]
\caption{Ablation study on rank of spurious correlation subspace and number of intervention layers. Evaluation metrics are the frame-level AUC, AP, and EER, respectively.}
\label{tab:different params}
\centering
\setlength{\tabcolsep}{8pt}
\renewcommand{\arraystretch}{1}

\resizebox{\textwidth}{!}{%
\begin{tabular}{c|c| ccc| ccc| ccc| ccc| ccc}
\toprule
\multirow{2}{*}{rank} & \multirow{2}{*}{layers}
& \multicolumn{3}{c|}{CDF\mbox{-}v2}
& \multicolumn{3}{c|}{DFDC}
& \multicolumn{3}{c|}{DFD}
& \multicolumn{3}{c|}{MobSwap}
& \multicolumn{3}{c}{BleFace} \\
\cmidrule(lr){3-5}\cmidrule(lr){6-8}\cmidrule(lr){9-11}\cmidrule(lr){12-14}\cmidrule(lr){15-17}

& &
AUC $\uparrow$ & AP $\uparrow$ & EER $\downarrow$ &
AUC $\uparrow$ & AP $\uparrow$ & EER $\downarrow$ &
AUC $\uparrow$ & AP $\uparrow$ & EER $\downarrow$ &
AUC $\uparrow$ & AP $\uparrow$ & EER $\downarrow$ &
AUC $\uparrow$ & AP $\uparrow$ & EER $\downarrow$ \\
\midrule

28 & 8
& 0.845 & 0.915 & 0.239
& 0.836 & 0.869 & 0.244
& 0.898 & 0.987 & 0.173
& 0.851 & 0.972 & 0.230
& 0.770 & 0.777 & 0.306 \\
32 & 8
& 0.858 & 0.922 & 0.230
& 0.842 & 0.868 & 0.237
& 0.917 & 0.990 & 0.154
& 0.877 & 0.977 & 0.205
& 0.764 & 0.773 & 0.309 \\
36 & 8
& 0.891 & 0.940 & 0.197
& 0.840 & 0.873 & 0.242
& 0.918 & 0.990 & 0.152
& 0.878 & 0.976 & 0.200
& 0.789 & 0.791 & 0.291\\
\midrule
28 & 12
& 0.883 & 0.936 & 0.198
& \textbf{0.868} & \textbf{0.899} & \textbf{0.209}
& 0.918 & 0.990 & 0.152
& 0.896 & 0.981 & 0.188
& 0.771 & 0.785 & 0.304 \\
32 & 12
& \textbf{0.901} & \textbf{0.947} & \textbf{0.187}
& \underline{0.853} & \underline{0.883} & \underline{0.233}
& \underline{0.926} & \underline{0.991} & \underline{0.149}
& \textbf{0.913} & \textbf{0.984} & \textbf{0.163}
& \textbf{0.807} & \textbf{0.818} & \textbf{0.265} \\
36 & 12
& \textbf{0.901} & \textbf{0.947} & \textbf{0.187}
& 0.838 & 0.865 & 0.242
& \textbf{0.932} & \textbf{0.992} & \textbf{0.142}
& 0.870 & 0.975 & 0.206
& \underline{0.790} & \underline{0.797} &\underline{0.274} \\
\midrule
28 & 16
& \underline{0.896} & \underline{0.943} & \underline{0.189}
& 0.839 & 0.865 & 0.247
& 0.915 & 0.989 & 0.163
& \underline{0.894} & \underline{0.980} & \underline{0.179}
& 0.775 & 0.784 & 0.289 \\
32 & 16
& 0.858 & 0.928 & 0.216
& 0.813 & 0.846 & 0.261
& 0.884 & 0.985 & 0.186
& 0.862 & 0.973 & 0.214
& 0.783 & 0.783 & 0.292 \\
36 & 16
& 0.819 & 0.883 & 0.258
& 0.713 & 0.757 & 0.352
& 0.849 & 0.978 & 0.234
& 0.754 & 0.942 & 0.316
& 0.709 & 0.710 & 0.351 \\

\bottomrule
\end{tabular}
}
\end{table*}

\begin{table}[t]
  \caption{Influence of intervention on various  CLIP architectures. (\textbf{Frame-level AUC}).}
  \label{tab:different LLMs}
  \centering
  \setlength{\tabcolsep}{5.0pt}
  \renewcommand{\arraystretch}{1.10}

  \begin{tabular}{c|ccc}
    \toprule
    Method  & CDF-v2 & DFDC & DFD \\
    \midrule
    CLIP ViT-B/32 \cite{clip} & 0.662 & 0.662 & 0.764 \\
    +LROR &
      \makecell[c]{\textbf{0.788}\\ \textcolor{blue}{\footnotesize$(\uparrow\,12.6\%)$}} &
      \makecell[c]{\textbf{0.760}\\ \textcolor{blue}{\footnotesize$(\uparrow\,9.8\%)$}} &
      \makecell[c]{\textbf{0.818}\\ \textcolor{blue}{\footnotesize$(\uparrow\,5.4\%)$}} \\
    \midrule
    CLIP ViT-B/16 \cite{clip} & 0.721 & 0.693 & 0.795 \\
    +LROR &
      \makecell[c]{\textbf{0.841}\\ \textcolor{blue}{\footnotesize$(\uparrow\,12.0\%)$}} &
      \makecell[c]{\textbf{0.804}\\ \textcolor{blue}{\footnotesize$(\uparrow\,11.1\%)$}} &
      \makecell[c]{\textbf{0.887}\\ \textcolor{blue}{\footnotesize$(\uparrow\,9.2\%)$}} \\
    \midrule
    CLIP ViT-L/14 \cite{clip} & 0.763 & 0.742 & 0.828 \\
    +LROR &
      \makecell[c]{\textbf{0.901}\\ \textcolor{blue}{\footnotesize$(\uparrow\,13.8\%)$}} &
      \makecell[c]{\textbf{0.853}\\ \textcolor{blue}{\footnotesize$(\uparrow\,11.1\%)$}} &
      \makecell[c]{\textbf{0.926}\\ \textcolor{blue}{\footnotesize$(\uparrow\,9.8\%)$}} \\
    \bottomrule
  \end{tabular}
\end{table}

\noindent \textbf{Impact of Different Hyperparameters.} In this section, we investigate the impact of different ranks of the low-rank subspace and numbers of intervention layers on detection performance of our model. As shown in the Table \ref{tab:different params}, we can observe that: (1) For the rank of the spurious correlation space, if rank is too small, it fails to eliminate most spurious correlation factors, and if rank is too large, it may hinder the sufficient learning of forgery-related features. (2) Regarding the number of intervention layers, fewer intervention cannot completely decouple spurious correlation in the visual space and excessive intervention may destroy pre-trained knowledge of the CLIP \cite{clip}. Therefore, we select rank=32 and layers=12 as the hyperparameters for the model in this paper.

\begin{figure*}[t]
  \centering
  \includegraphics[width=\textwidth]{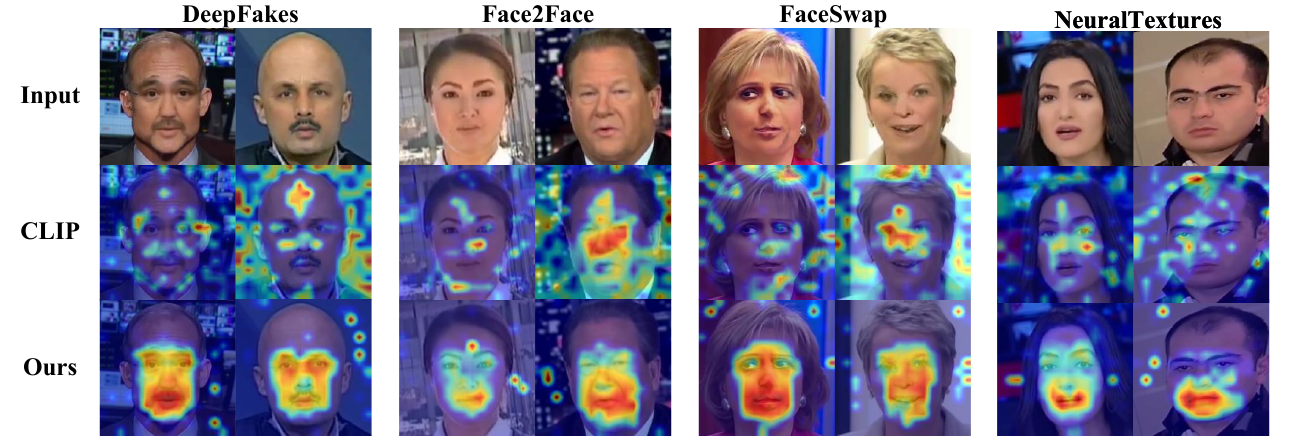}
  \caption{The visualization of GradCAM \cite{cam} on the Vanilla CLIP \cite{clip} and our method (\textbf{cross-manipulation}). The eight images in the top row correspond to fake images generated by the four forgery techniques, including DeepFake \cite{Deepfakes}, Face2Face \cite{thies2016face2face}, FaceSwap \cite{faceswap}, and NeuralTextures \cite{nt}. The following two rows show the attention maps for Vanilla CLIP and our method respectively.}
  \label{fig:gradcam1}
\end{figure*}

\begin{figure*}[t]
  \centering
  \includegraphics[width=\textwidth]{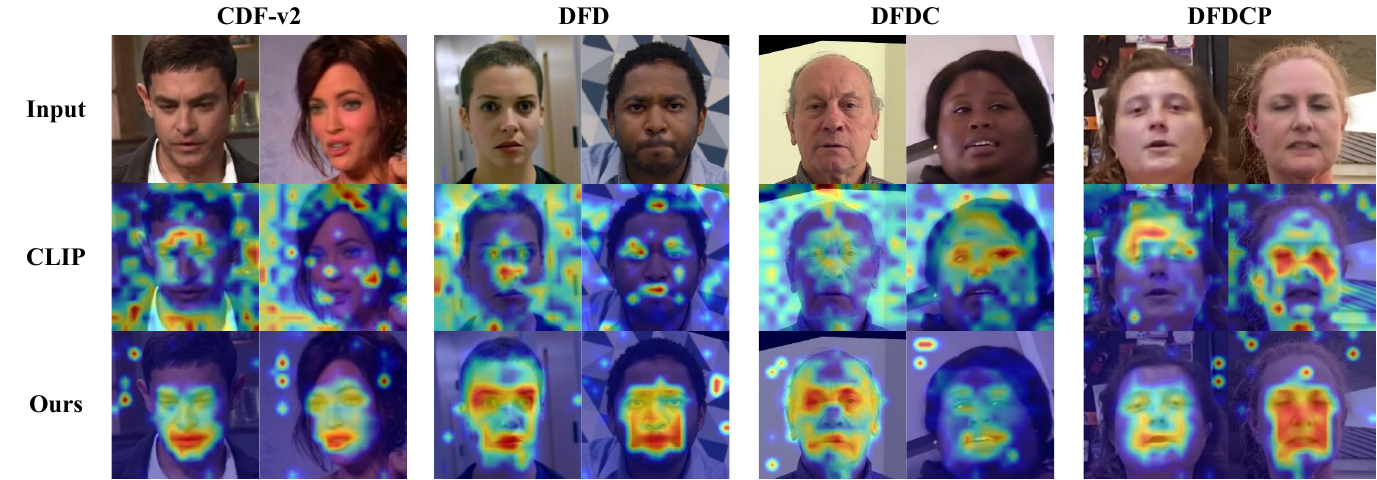}
  \caption{The visualization of GradCAM \cite{cam} on the Vanilla CLIP \cite{clip} and our method (\textbf{cross-datasets}). The eight images in the top row correspond to fake images from four standard benchmarks, including Celeb-DF-v2 \cite{li2020celeb}, DFD \cite{Dfd}, DFDC \cite{dfdc}, and DFDCP \cite{dfdcp}. The following two rows show the attention maps for Vanilla CLIP and our method respectively.}
  \label{fig:gradcam2}
\end{figure*}

\begin{figure}[t]
  \centering
  \includegraphics[width=0.85\columnwidth]{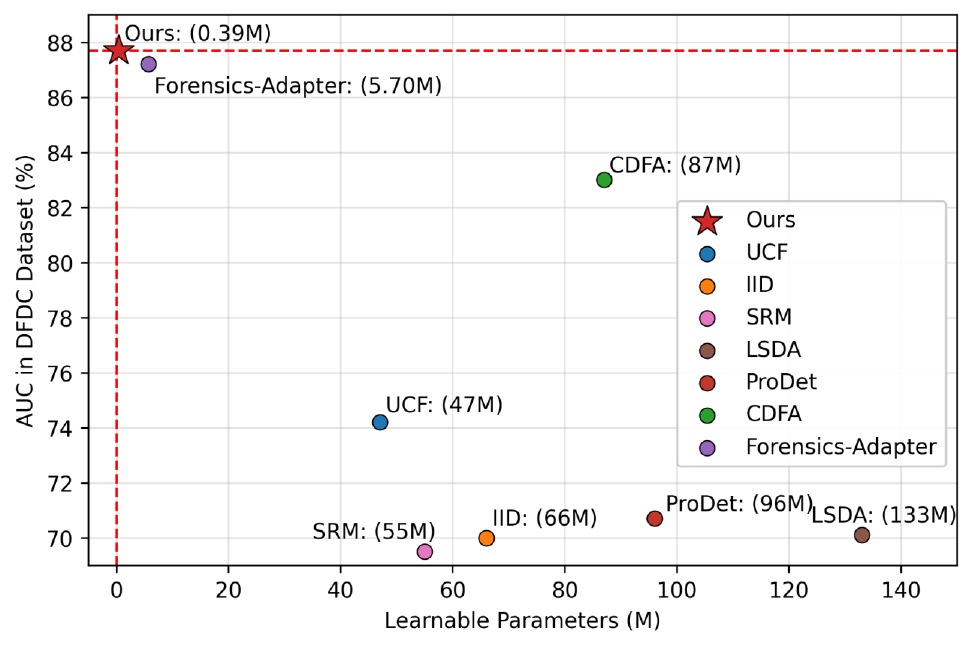}
  \caption{Comparison between our method and other SOTA detectors on the DFDC dataset (trained on FF++). Our method needs the fewest learnable parameters while achieves the best performance. }
  \label{fig:parameters}
\end{figure}

\noindent \textbf{Effects of intervention on different CLIP architectures.} In this section, we investigate the impact of visual-spatial interventions on different CLIP \cite{clip} architectures, including CLIP ViT-B/32, CLIP ViT-B/16, and CLIP ViT-L/14. Our approach utilize CLIP ViT-L/14 as backbone. As Table \ref{tab:different LLMs} clearly demonstrates, detection performance significantly improved across all CLIP variants following low-rank projection to eliminate spurious correlations. Notably, on the CDF-v2 dataset, all three architectures achieve over 12\% improvement in frame-level AUC performance after intervention. This fully validates the effectiveness of our methodology.

\noindent \textbf{Qualitative Results.} In this section, we use Grad-CAM \cite{cam} to visualize the location of model's attention on the faces generated by four forgery techniques (DeepFakes \cite{Deepfakes}, Face2Face \cite{thies2016face2face}, FaceSwap \cite{faceswap}, and NeuralTextures \cite{nt}) and four standard benchmarks (Celeb-DF-v2 \cite{li2020celeb}, DFD \cite{Dfd}, DFDC \cite{dfdc}, and DFDCP \cite{dfdcp}), which is shown in Figure \ref{fig:gradcam1}, \ref{fig:gradcam2}. It can be noticed that compared to baseline, our method better identifies and utilizes facial forgery cues to judge authenticity, greatly mitigating the negative interference of spurious correlation factors. Visualization results prove the effectiveness of our method.

\subsection{Complexity Analysis}

In this section, we examine the computational complexity of our approach. By freezing all parameters of CLIP and only training the low-rank orthogonal space for visual intervention alongside a linear classification head, our model requires a mere \textbf{0.39 million} trainable parameters. Compared to methods employing adapters for fine-tuning, our approach is remarkably lightweight. As demonstrated in Figure \ref{fig:parameters}, our approach achieves the best performance with the fewest trainable parameters, thereby proving its superiority.

\section{Conclusion}
In this paper, we analyze the reasons for Vanilla CLIP's generalization failure through GradCAM visualization and PCA empirical research: \textit{low-rank spurious bias}. To address this, we innovatively propose a simple and effective representation space intervention paradigm from the perspective of causal representation learning. Specifically, we remove the spurious correlation subspace through orthogonal low-rank projection, forcing the model to classify based on authentic causal factors. Both qualitative and quantitative analyses validate the effectiveness of the proposed method. Extensive experiments on several benchmark datasets demonstrate that our method possesses excellent generalization ability and robustness.

\bibliographystyle{IEEEtran}
\bibliography{reference.bib}

@inproceedings{rossler2019faceforensics++,
  title={Faceforensics++: Learning to detect manipulated facial images},
  author={Rossler, Andreas and Cozzolino, Davide and Verdoliva, Luisa and Riess, Christian and Thies, Justus and Nie{\ss}ner, Matthias},
  booktitle={Proceedings of the IEEE/CVF international conference on computer vision},
  pages={1--11},
  year={2019}
}

@misc{Deepfakes,
  title={Faceswapdevs},
  author={Deepfakes},
  howpublished={\url{https://github.com/ deepfakes/faceswap}},
  year={2019}
}

@inproceedings{thies2016face2face,
  title={Face2face: Real-time face capture and reenactment of rgb videos},
  author={Thies, Justus and Zollhofer, Michael and Stamminger, Marc and Theobalt, Christian and Nie{\ss}ner, Matthias},
  booktitle={Proceedings of the IEEE conference on computer vision and pattern recognition},
  pages={2387--2395},
  year={2016}
}

@misc{faceswap,
  title={Faceswap},
  author={Marek Kowalski},
  howpublished={\url{https://github.com/ deepfakes/faceswap}},
  year={2019}
}

@article{nt,
  title={Deferred neural rendering: Image synthesis using neural textures},
  author={Thies, Justus and Zollh{\"o}fer, Michael and Nie{\ss}ner, Matthias},
  journal={Acm Transactions on Graphics (TOG)},
  volume={38},
  number={4},
  pages={1--12},
  year={2019},
  publisher={ACM New York, NY, USA}
}

@inproceedings{li2020celeb,
  title={Celeb-df: A large-scale challenging dataset for deepfake forensics},
  author={Li, Yuezun and Yang, Xin and Sun, Pu and Qi, Honggang and Lyu, Siwei},
  booktitle={Proceedings of the IEEE/CVF conference on computer vision and pattern recognition},
  pages={3207--3216},
  year={2020}
}

@article{dfdcp,
  title={The deepfake detection challenge (dfdc) preview dataset. arXiv 2019},
  author={Dolhansky, B and Howes, R and Pflaum, B and Baram, N and Ferrer, CC},
  journal={arXiv preprint arXiv:1910.08854},
  year={2019}

}

@misc{Dfd,
  author       = {Deepfakedetection},
  howpublished = {\url{https://ai.googleblog.com/2019/09/contributing-data-to-deepfake-detection.html}},
  note         = {Accessed: 2021-11-13},
  year         = {2019}
}

@inproceedings{xception,
  title={Xception: Deep learning with depthwise separable convolutions},
  author={Chollet, Fran{\c{c}}ois},
  booktitle={Proceedings of the IEEE conference on computer vision and pattern recognition},
  pages={1251--1258},
  year={2017}
}

@inproceedings{efficientnet,
  title={Efficientnet: Rethinking model scaling for convolutional neural networks},
  author={Tan, Mingxing and Le, Quoc},
  booktitle={International conference on machine learning},
  pages={6105--6114},
  year={2019},
  organization={PMLR}
}

@inproceedings{f3,
  title={Thinking in frequency: Face forgery detection by mining frequency-aware clues},
  author={Qian, Yuyang and Yin, Guojun and Sheng, Lu and Chen, Zixuan and Shao, Jing},
  booktitle={European conference on computer vision},
  pages={86--103},
  year={2020},
  organization={Springer}
}

@inproceedings{recce,
  title={End-to-end reconstruction-classification learning for face forgery detection},
  author={Cao, Junyi and Ma, Chao and Yao, Taiping and Chen, Shen and Ding, Shouhong and Yang, Xiaokang},
  booktitle={Proceedings of the IEEE/CVF conference on computer vision and pattern recognition},
  pages={4113--4122},
  year={2022}
}

@inproceedings{sbi,
  title={Detecting deepfakes with self-blended images},
  author={Shiohara, Kaede and Yamasaki, Toshihiko},
  booktitle={Proceedings of the IEEE/CVF conference on computer vision and pattern recognition},
  pages={18720--18729},
  year={2022}
}

@article{f2trans,
  title={F 2 trans: High-frequency fine-grained transformer for face forgery detection},
  author={Miao, Changtao and Tan, Zichang and Chu, Qi and Liu, Huan and Hu, Honggang and Yu, Nenghai},
  journal={IEEE Transactions on Information Forensics and Security},
  volume={18},
  pages={1039--1051},
  year={2023},
  publisher={IEEE}
}

@inproceedings{ucf,
  title={Ucf: Uncovering common features for generalizable deepfake detection},
  author={Yan, Zhiyuan and Zhang, Yong and Fan, Yanbo and Wu, Baoyuan},
  booktitle={Proceedings of the IEEE/CVF International Conference on Computer Vision},
  pages={22412--22423},
  year={2023}
}

@article{cfm,
  title={Beyond the prior forgery knowledge: Mining critical clues for general face forgery detection},
  author={Luo, Anwei and Kong, Chenqi and Huang, Jiwu and Hu, Yongjian and Kang, Xiangui and Kot, Alex C},
  journal={IEEE Transactions on Information Forensics and Security},
  volume={19},
  pages={1168--1182},
  year={2023},
  publisher={IEEE}
}

@inproceedings{lsda,
  title={Transcending forgery specificity with latent space augmentation for generalizable deepfake detection},
  author={Yan, Zhiyuan and Luo, Yuhao and Lyu, Siwei and Liu, Qingshan and Wu, Baoyuan},
  booktitle={Proceedings of the IEEE/CVF Conference on Computer Vision and Pattern Recognition},
  pages={8984--8994},
  year={2024}
}

@inproceedings{caddm,
  title={Implicit identity leakage: The stumbling block to improving deepfake detection generalization},
  author={Dong, Shichao and Wang, Jin and Ji, Renhe and Liang, Jiajun and Fan, Haoqiang and Ge, Zheng},
  booktitle={Proceedings of the IEEE/CVF Conference on Computer Vision and Pattern Recognition},
  pages={3994--4004},
  year={2023}
}

@article{retinaface,
  title={Retinaface: Single-stage dense face localisation in the wild},
  author={Deng, Jiankang and Guo, Jia and Zhou, Yuxiang and Yu, Jinke and Kotsia, Irene and Zafeiriou, Stefanos},
  journal={arXiv preprint arXiv:1905.00641},
  year={2019}
}

@inproceedings{cam,
  title={Grad-cam: Visual explanations from deep networks via gradient-based localization},
  author={Selvaraju, Ramprasaath R and Cogswell, Michael and Das, Abhishek and Vedantam, Ramakrishna and Parikh, Devi and Batra, Dhruv},
  booktitle={Proceedings of the IEEE international conference on computer vision},
  pages={618--626},
  year={2017}
}

@inproceedings{FA,
  title={Forensics adapter: Adapting clip for generalizable face forgery detection},
  author={Cui, Xinjie and Li, Yuezun and Luo, Ao and Zhou, Jiaran and Dong, Junyu},
  booktitle={Proceedings of the Computer Vision and Pattern Recognition Conference},
  pages={19207--19217},
  year={2025}
}

@inproceedings{bi,
  title={Face x-ray for more general face forgery detection},
  author={Li, Lingzhi and Bao, Jianmin and Zhang, Ting and Yang, Hao and Chen, Dong and Wen, Fang and Guo, Baining},
  booktitle={Proceedings of the IEEE/CVF conference on computer vision and pattern recognition},
  pages={5001--5010},
  year={2020}
}

@inproceedings{srm,
  title={Generalizing face forgery detection with high-frequency features},
  author={Luo, Yuchen and Zhang, Yong and Yan, Junchi and Liu, Wei},
  booktitle={Proceedings of the IEEE/CVF conference on computer vision and pattern recognition},
  pages={16317--16326},
  year={2021}
}

@inproceedings{spsl,
  title={Spatial-phase shallow learning: rethinking face forgery detection in frequency domain},
  author={Liu, Honggu and Li, Xiaodan and Zhou, Wenbo and Chen, Yuefeng and He, Yuan and Xue, Hui and Zhang, Weiming and Yu, Nenghai},
  booktitle={Proceedings of the IEEE/CVF conference on computer vision and pattern recognition},
  pages={772--781},
  year={2021}
}

@inproceedings{core,
  title={Core: Consistent representation learning for face forgery detection},
  author={Ni, Yunsheng and Meng, Depu and Yu, Changqian and Quan, Chengbin and Ren, Dongchun and Zhao, Youjian},
  booktitle={Proceedings of the IEEE/CVF conference on computer vision and pattern recognition},
  pages={12--21},
  year={2022}
}

@inproceedings{ed,
  title={Exposing the deception: Uncovering more forgery clues for deepfake detection},
  author={Ba, Zhongjie and Liu, Qingyu and Liu, Zhenguang and Wu, Shuang and Lin, Feng and Lu, Li and Ren, Kui},
  booktitle={Proceedings of the AAAI Conference on Artificial Intelligence},
  volume={38},
  number={2},
  pages={719--728},
  year={2024}
}

@inproceedings{cdfa,
  title={Fake it till you make it: Curricular dynamic forgery augmentations towards general deepfake detection},
  author={Lin, Yuzhen and Song, Wentang and Li, Bin and Li, Yuezun and Ni, Jiangqun and Chen, Han and Li, Qiushi},
  booktitle={European Conference on Computer Vision},
  pages={104--122},
  year={2024},
  organization={Springer}
}

@article{beijing1,
  title={Face. evolve: A high-performance face recognition library},
  author={Wang, Qingzhong and Zhang, Pengfei and Xiong, Haoyi and Zhao, Jian},
  journal={arXiv preprint arXiv:2107.08621},
  year={2021}
}

@article{deepfakebench,
  title={Deepfakebench: A comprehensive benchmark of deepfake detection},
  author={Yan, Zhiyuan and Zhang, Yong and Yuan, Xinhang and Lyu, Siwei and Wu, Baoyuan},
  journal={arXiv preprint arXiv:2307.01426},
  year={2023}
}

@article{df40,
  title={Df40: Toward next-generation deepfake detection},
  author={Yan, Zhiyuan and Yao, Taiping and Chen, Shen and Zhao, Yandan and Fu, Xinghe and Zhu, Junwei and Luo, Donghao and Wang, Chengjie and Ding, Shouhong and Wu, Yunsheng and others},
  journal={arXiv preprint arXiv:2406.13495},
  year={2024}
}

@article{orthogonal,
  title={Orthogonal Subspace Decomposition for Generalizable AI-Generated Image Detection},
  author={Yan, Zhiyuan and Wang, Jiangming and Jin, Peng and Zhang, Ke-Yue and Liu, Chengchun and Chen, Shen and Yao, Taiping and Ding, Shouhong and Wu, Baoyuan and Yuan, Li},
  journal={arXiv preprint arXiv:2411.15633},
  year={2024}
}

@article{dfdc,
  title={The deepfake detection challenge (dfdc) dataset},
  author={Dolhansky, Brian and Bitton, Joanna and Pflaum, Ben and Lu, Jikuo and Howes, Russ and Wang, Menglin and Ferrer, Cristian Canton},
  journal={arXiv preprint arXiv:2006.07397},
  year={2020}
}

@inproceedings{robustness,
  title={Deeperforensics-1.0: A large-scale dataset for real-world face forgery detection},
  author={Jiang, Liming and Li, Ren and Wu, Wayne and Qian, Chen and Loy, Chen Change},
  booktitle={Proceedings of the IEEE/CVF conference on computer vision and pattern recognition},
  pages={2889--2898},
  year={2020}
}

@inproceedings{e4s,
  title={Fine-grained face swapping via regional gan inversion},
  author={Liu, Zhian and Li, Maomao and Zhang, Yong and Wang, Cairong and Zhang, Qi and Wang, Jue and Nie, Yongwei},
  booktitle={Proceedings of the IEEE/CVF conference on computer vision and pattern recognition},
  pages={8578--8587},
  year={2023}
}

@inproceedings{facedancer,
  title={Facedancer: Pose-and occlusion-aware high fidelity face swapping},
  author={Rosberg, Felix and Aksoy, Eren Erdal and Alonso-Fernandez, Fernando and Englund, Cristofer},
  booktitle={Proceedings of the IEEE/CVF winter conference on applications of computer vision},
  pages={3454--3463},
  year={2023}
}

@article{gaze,
  title={Where deepfakes gaze at? spatial-temporal gaze inconsistency analysis for video face forgery detection},
  author={Peng, Chunlei and Miao, Zimin and Liu, Decheng and Wang, Nannan and Hu, Ruimin and Gao, Xinbo},
  journal={IEEE Transactions on Information Forensics and Security},
  year={2024},
  publisher={IEEE}
}

@inproceedings{udd,
  title={Exploring unbiased deepfake detection via token-level shuffling and mixing},
  author={Fu, Xinghe and Yan, Zhiyuan and Yao, Taiping and Chen, Shen and Li, Xi},
  booktitle={Proceedings of the AAAI Conference on Artificial Intelligence},
  volume={39},
  number={3},
  pages={3040--3048},
  year={2025}
}

@article{FIAUSA,
  title={From specificity to generality: Revisiting generalizable artifacts in detecting face deepfakes},
  author={Ma, Long and Yan, Zhiyuan and Chen, Yize and Xu, Jin and Guo, Qinglang and Huang, Hu and Liao, Yong and Lin, Hui},
  journal={arXiv preprint arXiv:2504.04827},
  year={2025}
}

@article{GMDF,
  title={Gm-df: Generalized multi-scenario deepfake detection},
  author={Lai, Yingxin and Yu, Zitong and Yang, Jing and Li, Bin and Kang, Xiangui and Shen, Linlin},
  journal={arXiv preprint arXiv:2406.20078},
  year={2024}
}

@article{Moeffd,
  title={Moe-ffd: Mixture of experts for generalized and parameter-efficient face forgery detection},
  author={Kong, Chenqi and Luo, Anwei and Bao, Peijun and Yu, Yi and Li, Haoliang and Zheng, Zengwei and Wang, Shiqi and Kot, Alex C},
  journal={IEEE Transactions on Dependable and Secure Computing},
  year={2025},
  publisher={IEEE}
}

@inproceedings{realforensics,
  title={Leveraging real talking faces via self-supervision for robust forgery detection},
  author={Haliassos, Alexandros and Mira, Rodrigo and Petridis, Stavros and Pantic, Maja},
  booktitle={Proceedings of the IEEE/CVF conference on computer vision and pattern recognition},
  pages={14950--14962},
  year={2022}
}

@inproceedings{aunet,
  title={Aunet: Learning relations between action units for face forgery detection},
  author={Bai, Weiming and Liu, Yufan and Zhang, Zhipeng and Li, Bing and Hu, Weiming},
  booktitle={Proceedings of the IEEE/CVF conference on computer vision and pattern recognition},
  pages={24709--24719},
  year={2023}
}

@inproceedings{seeable,
  title={Seeable: Soft discrepancies and bounded contrastive learning for exposing deepfakes},
  author={Larue, Nicolas and Vu, Ngoc-Son and Struc, Vitomir and Peer, Peter and Christophides, Vassilis},
  booktitle={Proceedings of the IEEE/CVF International Conference on Computer Vision},
  pages={21011--21021},
  year={2023}
}

@inproceedings{iid,
  title={Implicit identity driven deepfake face swapping detection},
  author={Huang, Baojin and Wang, Zhongyuan and Yang, Jifan and Ai, Jiaxin and Zou, Qin and Wang, Qian and Ye, Dengpan},
  booktitle={Proceedings of the IEEE/CVF conference on computer vision and pattern recognition},
  pages={4490--4499},
  year={2023}
}

@inproceedings{laa,
  title={Laa-net: Localized artifact attention network for quality-agnostic and generalizable deepfake detection},
  author={Nguyen, Dat and Mejri, Nesryne and Singh, Inder Pal and Kuleshova, Polina and Astrid, Marcella and Kacem, Anis and Ghorbel, Enjie and Aouada, Djamila},
  booktitle={Proceedings of the IEEE/CVF Conference on Computer Vision and Pattern Recognition},
  pages={17395--17405},
  year={2024}
}

@article{talljiajia,
  title={Learning spatiotemporal inconsistency via thumbnail layout for face deepfake detection},
  author={Xu, Yuting and Liang, Jian and Sheng, Lijun and Zhang, Xiao-Yu},
  journal={International Journal of Computer Vision},
  volume={132},
  number={12},
  pages={5663--5680},
  year={2024},
  publisher={Springer}
}

@article{ffaa,
  title={Ffaa: Multimodal large language model based explainable open-world face forgery analysis assistant},
  author={Huang, Zhengchao and Xia, Bin and Lin, Zicheng and Mou, Zhun and Yang, Wenming and Jia, Jiaya},
  journal={arXiv preprint arXiv:2408.10072},
  year={2024}
}

@inproceedings{clip,
  title={Learning transferable visual models from natural language supervision},
  author={Radford, Alec and Kim, Jong Wook and Hallacy, Chris and Ramesh, Aditya and Goh, Gabriel and Agarwal, Sandhini and Sastry, Girish and Askell, Amanda and Mishkin, Pamela and Clark, Jack and others},
  booktitle={International conference on machine learning},
  pages={8748--8763},
  year={2021},
  organization={PmLR}
}

@inproceedings{FFD-STA,
  title={Generalizing deepfake video detection with plug-and-play: Video-level blending and spatiotemporal adapter tuning},
  author={Yan, Zhiyuan and Zhao, Yandan and Chen, Shen and Guo, Mingyi and Fu, Xinghe and Yao, Taiping and Ding, Shouhong and Wu, Yunsheng and Yuan, Li},
  booktitle={Proceedings of the Computer Vision and Pattern Recognition Conference},
  pages={12615--12625},
  year={2025}
}

@inproceedings{fcg,
  title={Towards More General Video-based Deepfake Detection through Facial Component Guided Adaptation for Foundation Model},
  author={Han, Yue-Hua and Huang, Tai-Ming and Hua, Kai-Lung and Chen, Jun-Cheng},
  booktitle={Proceedings of the Computer Vision and Pattern Recognition Conference},
  pages={22995--23005},
  year={2025}
}

@article{mfclip,
  title={MFCLIP: Multi-modal fine-grained CLIP for generalizable diffusion face forgery detection},
  author={Zhang, Yaning and Wang, Tianyi and Yu, Zitong and Gao, Zan and Shen, Linlin and Chen, Shengyong},
  journal={IEEE Transactions on Information Forensics and Security},
  year={2025},
  publisher={IEEE}
}

@inproceedings{gaints,
  title={Standing on the shoulders of giants: Reprogramming visual-language model for general deepfake detection},
  author={Lin, Kaiqing and Lin, Yuzhen and Li, Weixiang and Yao, Taiping and Li, Bin},
  booktitle={Proceedings of the AAAI Conference on Artificial Intelligence},
  volume={39},
  number={5},
  pages={5262--5270},
  year={2025}
}

@article{ProDet,
  title={Can we leave deepfake data behind in training deepfake detector?},
  author={Cheng, Jikang and Yan, Zhiyuan and Zhang, Ying and Luo, Yuhao and Wang, Zhongyuan and Li, Chen},
  journal={Advances in Neural Information Processing Systems},
  volume={37},
  pages={21979--21998},
  year={2024}
}

@inproceedings{lips,
  title={Lips don't lie: A generalisable and robust approach to face forgery detection},
  author={Haliassos, Alexandros and Vougioukas, Konstantinos and Petridis, Stavros and Pantic, Maja},
  booktitle={Proceedings of the IEEE/CVF conference on computer vision and pattern recognition},
  pages={5039--5049},
  year={2021}
}

@article{causal,
  title={Representation learning for treatment effect estimation from observational data},
  author={Yao, Liuyi and Li, Sheng and Li, Yaliang and Huai, Mengdi and Gao, Jing and Zhang, Aidong},
  journal={Advances in neural information processing systems},
  volume={31},
  year={2018}
}

@article{tsne,
  title={Visualizing data using t-SNE},
  author={Maaten, Laurens van der and Hinton, Geoffrey},
  journal={Journal of machine learning research},
  volume={9},
  number={Nov},
  pages={2579--2605},
  year={2008}
}

@book{causality,
  title={Causality},
  author={Pearl, Judea},
  year={2009},
  publisher={Cambridge university press}
}

@article{yaoqiu1,
  title={Local region frequency guided dynamic inconsistency network for deepfake video detection},
  author={Yue, Pengfei and Chen, Beijing and Fu, Zhangjie},
  journal={Big Data Mining and Analytics},
  volume={7},
  number={3},
  pages={889--904},
  year={2024},
  publisher={TUP}
}

@article{yaoqiu2,
  title={IIN-FFD: intra-inter network for face forgery detection},
  author={Zhou, Qihua and Zhou, Zhili and Bao, Zhipeng and Niu, Weina and Liu, Yuling},
  journal={Tsinghua Science and Technology},
  volume={29},
  number={6},
  pages={1839--1850},
  year={2024},
  publisher={TUP}
}

@article{CRL,
  title={Toward causal representation learning},
  author={Sch{\"o}lkopf, Bernhard and Locatello, Francesco and Bauer, Stefan and Ke, Nan Rosemary and Kalchbrenner, Nal and Goyal, Anirudh and Bengio, Yoshua},
  journal={Proceedings of the IEEE},
  volume={109},
  number={5},
  pages={612--634},
  year={2021},
  publisher={IEEE}
}

@article{causalpaper1,
  title={Improving deepfake detection generalization by invariant risk minimization},
  author={Yin, Zixin and Wang, Jiakai and Xiao, Yisong and Zhao, Hanqing and Li, Tianlin and Zhou, Wenbo and Liu, Aishan and Liu, Xianglong},
  journal={IEEE Transactions on Multimedia},
  volume={26},
  pages={6785--6798},
  year={2024},
  publisher={IEEE}
}

@article{causalclip-causalpaper2,
  title={CausalCLIP: Causally-Informed Feature Disentanglement and Filtering for Generalizable Detection of Generated Images},
  author={Liu, Bo and Qin, Qiao and He, Qinghui},
  journal={arXiv preprint arXiv:2512.13285},
  year={2025}
}

@inproceedings{tall,
  title={Tall: Thumbnail layout for deepfake video detection},
  author={Xu, Yuting and Liang, Jian and Jia, Gengyun and Yang, Ziming and Zhang, Yanhao and He, Ran},
  booktitle={Proceedings of the IEEE/CVF international conference on computer vision},
  pages={22658--22668},
  year={2023}
}

@inproceedings{FTCNTTN,
  title={Exploring temporal coherence for more general video face forgery detection},
  author={Zheng, Yinglin and Bao, Jianmin and Chen, Dong and Zeng, Ming and Wen, Fang},
  booktitle={Proceedings of the IEEE/CVF international conference on computer vision},
  pages={15044--15054},
  year={2021}
}

@article{LPSssd,
  title={Deepfake video detection via predictive representation learning},
  author={Ge, Shiming and Lin, Fanzhao and Li, Chenyu and Zhang, Daichi and Wang, Weiping and Zeng, Dan},
  journal={ACM Transactions on Multimedia Computing, Communications, and Applications (TOMM)},
  volume={18},
  number={2s},
  pages={1--21},
  year={2022},
  publisher={ACM New York, NY}
}

@inproceedings{two-branch,
  title={Two-branch recurrent network for isolating deepfakes in videos},
  author={Masi, Iacopo and Killekar, Aditya and Mascarenhas, Royston Marian and Gurudatt, Shenoy Pratik and AbdAlmageed, Wael},
  booktitle={European conference on computer vision},
  pages={667--684},
  year={2020},
  organization={Springer}
}

@inproceedings{TD3DCNN,
  title={Detecting Deepfake Videos with Temporal Dropout 3DCNN.},
  author={Zhang, Daichi and Li, Chenyu and Lin, Fanzhao and Zeng, Dan and Ge, Shiming},
  booktitle={IJCAI},
  pages={1288--1294},
  year={2021}
}

@inproceedings{PEL,
  title={Exploiting fine-grained face forgery clues via progressive enhancement learning},
  author={Gu, Qiqi and Chen, Shen and Yao, Taiping and Chen, Yang and Ding, Shouhong and Yi, Ran},
  booktitle={Proceedings of the AAAI conference on artificial intelligence},
  volume={36},
  number={1},
  pages={735--743},
  year={2022}
}

@inproceedings{STDT,
  title={Deepfake video detection with spatiotemporal dropout transformer},
  author={Zhang, Daichi and Lin, Fanzhao and Hua, Yingying and Wang, Pengju and Zeng, Dan and Ge, Shiming},
  booktitle={Proceedings of the 30th ACM international conference on multimedia},
  pages={5833--5841},
  year={2022}
}

@article{PCC,
  title={Learning patch-channel correspondence for interpretable face forgery detection},
  author={Hua, Yingying and Shi, Ruixin and Wang, Pengju and Ge, Shiming},
  journal={IEEE Transactions on Image Processing},
  volume={32},
  pages={1668--1680},
  year={2023},
  publisher={IEEE}
}

@inproceedings{MADD,
  title={Multi-attentional deepfake detection},
  author={Zhao, Hanqing and Zhou, Wenbo and Chen, Dongdong and Wei, Tianyi and Zhang, Weiming and Yu, Nenghai},
  booktitle={Proceedings of the IEEE/CVF conference on computer vision and pattern recognition},
  pages={2185--2194},
  year={2021}
}

@article{mapmamba,
  title={MAP-Mamba: Multi-Artifacts Perception Mamba for Generalizable Face Forgery Detection},
  author={Wang, Chi and He, Ziwen and Hu, Xinjue and Guan, Weinan and Wang, Wei and Fu, Zhangjie},
  journal={IEEE Transactions on Information Forensics and Security},
  year={2026},
  publisher={IEEE}
}

@inproceedings{rfm,
  title={Representative forgery mining for fake face detection},
  author={Wang, Chengrui and Deng, Weihong},
  booktitle={Proceedings of the IEEE/CVF conference on computer vision and pattern recognition},
  pages={14923--14932},
  year={2021}
}

@article{ddl,
  title={Ddl: A dataset for interpretable deepfake detection and localization in real-world scenarios},
  author={Miao, Changtao and Zhang, Yi and Gao, Weize and Luo, Man and Feng, Weiwei and Tan, Zhiya and Li, Jianshu and Liu, Ajian and Diao, Yunfeng and Chu, Qi and others},
  journal={arXiv e-prints},
  pages={arXiv--2506},
  year={2025}
}

\vfill

\end{document}